\documentclass[letterpaper]{article} 
\usepackage[]{arxiv}  
\usepackage{times}  
\usepackage{helvet}  
\usepackage{courier}  
\usepackage[hyphens]{url}  
\usepackage{graphicx} 
\usepackage{natbib}  
\usepackage{caption} 
\usepackage{algorithm}
\usepackage{algpseudocode}

\usepackage{newfloat}
\usepackage{listings}
\DeclareCaptionStyle{ruled}{labelfont=normalfont,labelsep=colon,strut=off} 
\floatstyle{ruled}
\newfloat{listing}{tb}{lst}{}
\floatname{listing}{Listing}
\usepackage[export]{adjustbox}
\usepackage[utf8]{inputenc} 
\usepackage[T1]{fontenc}    

\usepackage{url}            
\usepackage{booktabs}       
\usepackage{amsfonts}       
\usepackage{nicefrac}       
\usepackage{microtype}      
\usepackage[dvipsnames]{xcolor}         
\usepackage{amsmath}
\usepackage{BOONDOX-cal}

\usepackage{tikz}           
\usetikzlibrary{arrows.meta} 
\usepackage{tikz-feynman}   
\definecolor{rwth-blue}{cmyk}{1,.5,0,0}
\definecolor{rwth-violet}{cmyk}{.6,.6,0,0}
\definecolor{rwth-purple}{cmyk}{.7,1,.35,.15}
\definecolor{rwth-carmine}{cmyk}{.25,1,.7,.2}
\definecolor{rwth-red}{cmyk}{.15,1,1,0}
\definecolor{rwth-magenta}{cmyk}{0,1,.25,0}
\definecolor{rwth-orange}{cmyk}{0,.4,1,0}
\definecolor{rwth-yellow}{cmyk}{0,0,1,0}
\definecolor{rwth-grass}{cmyk}{.35,0,1,0}
\definecolor{rwth-green}{cmyk}{.7,0,1,0}

\usepackage[colorlinks,linkcolor=NavyBlue,urlcolor=black,citecolor=ForestGreen]{hyperref}       

\usepackage{bm}

\usepackage{graphicx}
\usepackage{subcaption}
\usepackage{multirow}
\newcommand{\method}{\textsc{ANYCSP}}

\newcommand{\rb}{\textsc{Model RB}}
\newcommand{\mcut}{\textsc{MaxCut}}
\newcommand{\col}{\textsc{$k$\nobreakdash-Col}}
\newcommand{\msat}{\textsc{Max\nobreakdash-$k$\nobreakdash-SAT}}
\newcommand{\threesat}{\textsc{3-SAT}}
\newcommand{\csp}{\textsc{CSP}}

\usepackage[textwidth=1.7cm, disable]{todonotes}
\newcommand{\mnote}[3]{\todo[color=#3!40,size=\footnotesize]{\textbf{#2:} #1}}

\newcommand{\martin}[1]{\mnote{#1}{M}{blue}}

\newcommand{\jan}[1]{\mnote{#1}{J}{red}}

\newcommand{\CC}{{\mathcal C}}
\newcommand{\CD}{{\mathcal D}}
\newcommand{\CI}{{\mathcal I}}
\newcommand{\CV}{{\mathcal V}}
\newcommand{\CX}{{\mathcal X}}
\newcommand{\Cv}{{\mathcal v}}

\newcommand{\er}{Erd\H{o}s-R\'{e}nyi}
\newcommand{\ba}{Barab\'asi-Albert}

\newcommand{\citeay}[1]{\citeauthor{#1}~\citeyear{#1}}

\title{One Model, Any CSP: Graph Neural Networks as\\ Fast Global Search Heuristics for Constraint Satisfaction}

\iftrue
\title{One Model, Any CSP: Graph Neural Networks as\\ Fast Global Search Heuristics for Constraint Satisfaction}
\author {
    Jan Tönshoff\textsuperscript{\rm 1},
    Berke Kisin\textsuperscript{\rm 2},
    Jakob Lindner\textsuperscript{\rm 3},
    Martin Grohe\textsuperscript{\rm 4}
}
\affiliations {
    \textsuperscript{\rm 1} RWTH Aachen, toenshoff@informatik.rwth-aachen.de\\
    \textsuperscript{\rm 2} RWTH Aachen, berke.kisin@rwth-aachen.de\\
    \textsuperscript{\rm 3} RWTH Aachen, jakob.lindner@rwth-aachen.de\\
    \textsuperscript{\rm 4} RWTH Aachen, grohe@informatik.rwth-aachen.de\\
}
\fi

\begin{document}

\maketitle

\begin{abstract}
    We propose a universal Graph Neural Network architecture which can be trained as an end-2-end search heuristic for any Constraint Satisfaction Problem (CSP).
    Our architecture can be trained unsupervised with policy gradient descent to generate problem specific heuristics for any CSP in a purely data driven manner.
    The approach is based on a novel graph representation for CSPs that is both generic and compact and enables us to process every possible CSP instance with one GNN, regardless of constraint arity, relations or domain size.
    Unlike previous RL-based methods, we operate on a global search action space and allow our GNN to modify any number of variables in every step of the stochastic search.
    This enables our method to properly leverage the inherent parallelism of GNNs.
    We perform a thorough empirical evaluation where we learn heuristics for well known and important CSPs from random data, including graph coloring, \mcut{}, \threesat{} and \msat{}.
    Our approach outperforms prior approaches for neural combinatorial optimization by a substantial margin.
    It can compete with, and even improve upon, conventional search heuristics on test instances that are several orders of magnitude larger and structurally more complex than those seen during training.
\end{abstract}

\section{Introduction}
Constraint Satisfaction Problems (CSP) are a ubiquitous framework for specifying combinatorial search and optimization problems. They include many of the best-known NP-hard problems such as Boolean satisfiability (\textsc{Sat}), graph coloring (\textsc{Col}) and maximum cut (\mcut{}) 
and can flexibly adapted to model specific application dependent problems. CSP solution strategies range from general solvers based on methods such as constraint propagation or local search (see \citet{RussellN2020}, Chapter~6)
 to specialized solvers for individual problems like \textsc{Sat} (see \citet{BiereHMW2021}). In recent years, there is a growing interest in applying deep learning methods to combinatorial problems including many CSPs (e.g.~\citet{khalil2017learning}, \citet{selsam2018learning}, \citet{10.3389/frai.2020.580607}). The main motivation for these approaches is to learn novel heuristics from data rather than crafting them by hand.

Graph Neural Networks (see \citeay{gilmer2017neural}) have emerged as an effective tool for learning powerful, permutation invariant functions on graphs using deep neural networks, and they have become one of the main architectures in the field of neural combinatorial optimization. Problem instances are modelled as graphs and then mapped to approximate solutions with GNNs.
However, most methods use graph reductions and GNN architectures that are problem specific, and transferring them across combinatoral tasks requires considerable engineering, limiting their use cases.
\emph{Designing a generic neural network architecture and training procedure for the general CSP formalism offers a powerful alternative.} Then learning heuristics for any specific CSP becomes a purely data driven process requiring no specialized graph reduction or architecture search.

We propose a novel GNN based reinforcement learning approach to general constraint satisfaction. The main contributions of our method called \method{}\footnote{\textbf{A}re \textbf{N}eural Networks great heuristics? \textbf{Y}es, for \textbf{CSP}s.} can be summarized as follows:
We define a new graph representation for general CSP instances which is generic and well suited as an input for recurrent GNNs.
It allows us to directly process all CSPs with one unified architecture and no prior reduction to more specific CSPs, such as SAT.
In particular, one \method{} model can take every CSP instance as input, even those with domain sizes, constraint arities, or relations not seen during training.
Training is unsupervised using policy gradient ascent with a carefully tailored reward scheme that encourages exploration and prevents the search to get stuck in local maxima. 
During inference, a trained \method{} model iteratively searches the space of assignments to the variables of the CSP instance for an optimal solution satisfying the maximum number of constraints. 
Crucially, the search is \emph{global}; it allows transitions between any two assignments in a single step.
To enable this global search we use \emph{policy gradient} methods to handle the exponential action spaces efficiently. 
This design choice speeds up the search substantially, especially on large instances.
We thereby overcome a primary bottleneck of previous neural approaches based on local search, which only flips the values of a single or a few variables in each step. 
GNN based local search tends to scale poorly to large instances as one GNN forward pass takes significantly more time than one step of classical local search heuristics.
\method{} accounts for this and exploits the GNNs inherent parallelism to refine the solution globally in each step.

We evaluate \method{} by learning heuristics for a range of important CSPs: \textsc{Col}, \textsc{Sat}, \mcut{} and general CSP benchmark instances. 
We demonstrate that our method achieves a substantial increase in performance over prior GNN approaches and can compete with conventional algorithms.
\method{} models trained on small random graph coloring problems are on par with state-of-the-art coloring heuristics on structured benchmark instances. 
On \msat{}, our method scales to test instances 100 times larger than the training data, where it finds better assignments than state-of-the-art conventional search heuristics despite performing 1000 times fewer search iterations.
\section{Related Work}
For a comprehensive overview on applying GNNs to combinatorial problems, we refer to \citet{ijcai2021-595}.
In this paper, we are primarily interested in end-2-end approaches which seek to directly predict approximate solutions for combinatorial problems with trainable neural networks.
Early work in this area was done by \citet{bello2016neural}, who learned TSP heuristics with Pointer Networks \cite{vinyals2015pointer} and policy gradient descent.
Several extensions of these ideas have since been proposed based on attention \cite{kool2018attention} and GNNs \cite{DBLP:journals/corr/abs-2006-07054}.
\citet{khalil2017learning} propose a general method for graph problems, such as \mcut{} or Minimum Vertex Cover.
They model the expansion of partial solutions as a reinforcement learning task and train a GNN with Q-learning to iteratively construct approximate solutions.

A related group of approaches models local modifications to complete solutions as actions of a reinforcement learning problem.
A GNN is then trained as a local search heuristic that iteratively improves candidate solutions through local changes.
Methods following this concept are RLSAT \cite{yolcu2019learning} for SAT,  ECO-DQN \cite{barrett2020exploratory} for \mcut{}, LS-DQN \cite{yao2021reversible} for graph partitioning problems and TSP as well as BiHyb \cite{NEURIPS2021_b2f627ff} for graph problems based on selecting and modifying edges.
Like conventional search heuristics, these architectures can be applied for any number of search iterations to refine the solution.
A shared drawback on large instances is the relatively high computational cost of GNNs, which slows down the search substantially when compared to classical algorithms. 
ECORD \cite{barrett2022learning} addresses this issue for \mcut{} by applying a GNN only once before the local search, which is carried out by a faster GRU-based architecture without costly message passes.
We address the same problem, but not by iterating faster, but by allowing global modifications in each iteration.


A fundamentally different approach considers soft relaxations of the underlying problems which can optimized directly with SGD.
Examples of this concept are PDP \cite{amizadeh2019pdp} for SAT and RUNCSP \cite{10.3389/frai.2020.580607} for all binary CSPs with fixed constraint language.
These architectures can predict completely new solutions in each iteration but the relaxed differentiable objectives used for training typically do not capture the full hardness of the discrete problem.

\section{Preliminaries}


A \emph{CSP instance} is a triple $\CI=(\CX,\CD,\CC)$, where $\CX$ is a finite set of variables, $\CD$ assigns to each variable $X\in\CX$ a finite set $\CD(X)$, the \emph{domain} of $X$,  and
$\CC$ is a set of constraints $C=\big(s^C,R^C)$, where for some $k\ge 1$, the \emph{scope} $s^C=(X_1,\ldots,X_k)\in\CX^k$ is a tuple of variables and $R^C\subseteq\CD(X_1)\times\ldots\times\CD(X_k)$ is a $k$-ary relation over the corresponding domains. We always assume that the variables in the scope $s^C$ of a constraint $C$ are mutually distinct; we can easily transform an instance not satisfying this condition into one that does by adapting the relation $R^C$ accordingly.

Slightly abusing terminology, we call a pair $(X,d)$ where $X\in\CX$ and $d\in\CD(X)$ a \emph{value} for variable $X$.
For all $X\in\CX$ we let $\CV_X=\{X\}\times\CD(X)$ be the set of all values for $X$, and we let $\CV=\bigcup_{X\in\mathcal{X}}\CV_X$ be the set of all values. We usually denote values by $\Cv$. Working with these values instead of domain elements is convenient because the sets $\CV_X$, for $X\in\CX$, are mutually disjoint, whereas the domains $\CD(X)$ are not necessarily.

An \emph{assignment} for a CSP instance $\mathcal{I}=(\CX,\CD,\CC)$ is a mapping $\alpha$ that assigns a domain element $\alpha(X)\in\CD(X)$ to each variable $X$. Alternatively, we may view an assignment as a subset $\alpha\subseteq\CV$ that contains exactly one value from each $\CV_X$. Depending on the context, we use either view, and we synonymously write $\alpha(X)=d$ or $(X,d)\in\alpha$. An assignment $\alpha$ \emph{satisfies} a constraint $C=((X_1,\ldots,X_k),R)$ (we write $\alpha\models C$) if $(\alpha(X_1),\ldots,\alpha(X_k))\in R$, and $\alpha$ \emph{satisfies} $\CI$, or is a \emph{solution} to $\CI$, if it satisfies all constraints in $\CC$.
The objective of a \textsc{CSP} is to decide if a given instance has a satisfying assignment and to find one if it does. To distinguish this problem from the maximization version introduced below, we sometimes speak of the \emph{decision version}. Specific CSPs such as Boolean satisfiability or graph coloring problems are obtained by restricting the instances considered.

We define the \emph{quality} $Q_\CI(\alpha)$ of an assignment $\alpha$ to be the fraction
of constraints in $\CC$ satisfied by $\alpha$:
$
    Q_\mathcal{I}(\alpha) = |\{C | C \in \mathcal{C}, \alpha \models C\}|/|\mathcal{C}|.
$
An assignment $\alpha$ is \emph{optimal} if it maximizes $Q_\mathcal{I}(\alpha)$ for the instance $\mathcal{I}$. The goal of the maximisation problem \textsc{MaxCSP} is to find an optimal assignment for a given instance.

A \emph{soft assignment} for a CSP instance $\CI$ is a mapping $\varphi:\CV\to[0,1]$  such that $\sum_{\Cv \in\CV_X} \varphi(\Cv) = 1$ for all $X\in\CX$. We interpret the numbers $\varphi(\Cv)$ as probabilities and say that an assignment $\alpha$ is \emph{sampled} from a soft assignment $\varphi$ (we write $\alpha\sim\varphi$) if for each variable $X\in\mathcal{X}$ we independently draw a value $\Cv \in\CV_X$ with probability $\varphi(\Cv)$.


\section{Method}
\begin{figure*}[t]
        \centering
        \noindent\begin{tikzpicture}[
every label/.style = {fill=white, rectangle, draw=black!50, rounded corners=3pt, minimum width=5pt, shift={(-0.0,-0.67)}, inner sep= 3pt},
nodes={
draw,
circle,
minimum width=21pt, 
line width=2pt,
inner sep=1pt}
]
    \node at (0,0) [fill= rwth-green!25, draw= rwth-green!70] (X1) {$X$};
    \node at (3, 0) [fill= rwth-green!25, draw= rwth-green!70] (X2)  {$Y$};
    \node at (5.45, 0) [fill= rwth-green!25, draw= rwth-green!70] (X3)  {$Z$};
    \draw[line width=1.5pt] (X1) -- (0,-1.1);
    \draw[line width=1.5pt] (X3) -- (4.9, -1.1);
    \draw[line width=1.5pt] (X3) -- (6, -1.1);
    \draw[line width=1.5pt] (X1) -- (-1.1, -1.1);
    \draw[line width=1.5pt] (X1) -- (1.1, -1.1);
    \draw[line width=1.5pt] (X2) -- (2.45, -1.1);
    \draw[line width=1.5pt] (X2) -- (3.55, -1.1);
    \node at (-1.1, -1.1) [fill= rwth-blue!25, draw= rwth-blue!70, label={0}] (1)  {}; 
    \node at (0,-1.1) [fill= rwth-blue!25, draw= rwth-blue!70, label={1}] (2)  {}; 
    \node at (1.1, -1.1) [fill= rwth-blue!25, draw= rwth-blue!70, label={0}] (3)  {};
    \node at (2.45, -1.1) [fill= rwth-blue!25, draw= rwth-blue!70, label={1}] (4)  {}; 
    \node at (3.55, -1.1) [fill= rwth-blue!25, draw= rwth-blue!70, label={0}] (5)  {}; 
    \node at (4.9, -1.1) [fill= rwth-blue!25, draw= rwth-blue!70, label={0}] (6)  {}; 
    \node at (6, -1.1) [fill= rwth-blue!25, draw= rwth-blue!70, label={1}] (7)  {};
    \node at (1.5, -2.8) [fill= rwth-carmine!25, draw= rwth-carmine!70] (C1)  {$C_1$};
    \node at (4.225, -2.8) [fill= rwth-carmine!25, draw= rwth-carmine!70] (C2)  {$C_2$}; 
    \path[every node/.style={fill=white, rectangle, draw=black!50, rounded corners=3pt}, line width=1.5pt]
        (C1) edge  node{1} (1)            
        (C1) edge  node{0} (2)
        (C1) edge  node{0} (3)
        (C1) edge  node{0} (4)
        (C1) edge  node{1} (5)
        (C2) edge  node{1} (4)            
        (C2) edge  node{0} (5)
        (C2) edge  node{0} (6)
        (C2) edge  node{1} (7);
    \node at (2.45, 0.8) [draw=none,rectangle,] {$G(\mathcal{I},\alpha):$};
    \node at (-3.6, -1.1) [draw=none, rectangle, align=flush left]
    {
    CSP Instance $\mathcal{I}:$ \\[0.7em]
    \textcolor{ForestGreen}{$\mathcal{X} = \{X,Y,Z\}$} \\[0.7em]
     \textcolor{rwth-blue}{$\CD(X) = \{1,2,3\}$} \\
    \textcolor{rwth-blue}{ $\CD(Y) = \{1,2\}$} \\
     \textcolor{rwth-blue}{$\CD(Z) = \{1,2\}$} \\[0.7em]
     \textcolor{rwth-carmine}{$C_1:X \leq Y$} \\
    \textcolor{rwth-carmine}{ $C_2: Y \neq Z$}\\[0.7em]
    Assignment $\alpha =(2,1,2)$ 
    };
\end{tikzpicture}
        \caption{Example of the \textbf{constraint value graph} $G(\mathcal{I},\alpha)$ for a given CSP instance $\mathcal{I}$ and an assignment $\alpha$.
        The graph contains vertices for the variables, values and constraints of $\mathcal{I}$.
        Each value is connected to its variable and labeled with the assignment $\alpha$.
        Each constraint is connected to the values of its variables.
        This edge set is labeled such that a label of $1$ for edge $(C,\Cv)$ states that choosing value $v$ will satisfy the constraint $C$ if no other variables involved in $C$ change their values.} 
        \label{Fig:Graph}
\end{figure*}
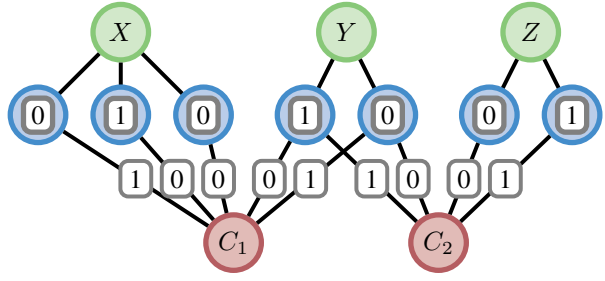
With every CSP instance $\CI=(\CX,\CD,\CC)$ we associate a tripartite graph 
with vertex set $\CX\cup\CV\cup\CC$, where $\CV$ is the set of values defined in the previous section, and two kinds of edges: \emph{variable edges} $(X,\Cv)$ for all $X\in\CX$ and $\Cv\in\CV_X$, and \emph{constraint edges} $(C,\Cv)$ for all $C\in\CC$ and $\Cv\in\CV_X$ for some $X$ in the scope of $C$. 

This graph representation is more or less standard; one slightly unusual feature is that we introduce edges from constraints directly to the values and not to the variables. 
This will be important in the next step, where information about the constraint relations $R^C$ is compactly encoded through a binary labeling of the constraint edges. 
For each assignment $\alpha$ we introduce a vertex labeling $L_V$  and an edge labeling $L_E$. 
The vertex labeling $L_V$ is a binary encoding of $\alpha$, that is, $L_V(\Cv)=1$ if $\Cv\in\alpha$ and $L_V(\Cv)=0$ for each $\Cv\in\CV\setminus\alpha$. 
The edge labeling $L_E$ 
encodes how changes to $\alpha$ affect each constraint.
For every constraint $C\in\CC$ and value $(X_i,d) \in \CV_{X_i}$ of variable $X_i$ in the scope $(X_1,\dots,X_k)$ of $C$ we define the edge label to be $L_E(C,\Cv)=1$ if 
\begin{equation*}
    (\alpha(X_1),\dots,\alpha(X_{i-1}),d,\alpha(X_{i+1}),\dots,\alpha(X_{k})) \in R^C.
\end{equation*}
and $L_E(C,\Cv)=0$ otherwise. 
Intuitively, the edge labels encode for each constraint edge $(C,\Cv)$ whether or not choosing the value $\Cv$ for its variable would satisfy $C$ under the condition that all other variables involved in $C$ retain their current value in $\alpha$.
We call the labeled graph $G(\CI,\alpha)$ obtained this way the \emph{constraint value graph} of $\CI$ at $\alpha$. 
Figure \ref{Fig:Graph} provides a visual example of our construction.

\subsection{Architecture}
\label{Section:Architecture}
We construct a recurrent GNN $\pi_\theta$ that maps constraint value graphs to soft assignments and serves as a trainable policy for our reinforcement-learning setup. 
Here, the real vector $\theta$ contains the trainable parameters of $\pi_\theta$.
The input of $\pi_\theta$ in iteration $t$ is the current graph $G(\CI,\alpha^{(t-1)})$ and recurrent vertex states $h^{(t-1)}$.
The output is a new soft assignment $\varphi^{(t)}$ for $\CI$ as well as updated recurrent states:
\begin{equation}
	\varphi^{(t)}, h^{(t)} = \pi_\theta\big(G(\mathcal{I}, \alpha^{(t-1)}), h^{(t-1)}\big)
\end{equation}
The next assignment $\alpha^{(t)}$ can then be sampled from  $\varphi^{(t)}$ before the process is repeated.
Here, we will provide an overview of the GNN architecture while we give a detailed formal description in Appendix A. 

In a nutshell, our architecture is a recurrent heterogeneous GNN that uses distinct trainable functions for each of the three vertex types in the constraint value graph.
The main hyperparameters of $\pi_\theta$ are the latent dimension $d\in\mathbb{N}$ and the aggregation function $\bigoplus$ which we either choose as an element-wise SUM, MEAN or MAX function. 
As a rule of thumb, we found MAX-aggregation to perform best on decision problems while MEAN-aggregation seems more suitable for maximization tasks.
This coincides with observations of \citet{DBLP:journals/corr/abs-2006-07054}.

$\pi_\theta$ associates a recurrent state $h^{(t)}(\Cv) \in \mathbb{R}^d$ with each value $\Cv \in \CV$ and uses a GRU cell to update these states after each round of message passing.
Variables and constraints do not have recurrent states.
We did consider versions with stateful constraints and variables, but these did not perform better while being slower. 
All remaining functions for message generation and combination are parameterized by standard MLPs with at most one hidden layer.
In each iteration $t$, $\pi_\theta$ performs 4 directed message passes in the following order:
(1) values to constraints, (2) constraints to values, (3) values to variables, (4) variables to values.
The first two message passes incorporate the node and edge labels and enable the values to gather information about how changes to the current assignment effect each constraint.
The final two message passes allow the values of each domain to negotiate the next variable assignment.
Note that this procedure is carried out \emph{once} in each search iteration $t$.
As the recurrent states can carry aggregated information across search iterations we found a single round of message passes per iteration sufficient.

Finally, $\pi_\theta$ generates a new soft assignment $\varphi^{(t)}$.
To this end, each value $\Cv \in \CV_X$ of each variable $X$ predicts a scalar real number $o^{(t)}(\Cv)=\mathbf{O}(h^{(t)}(\Cv))$ from its updated latent state with a shared MLP $\mathbf{O}:\mathbb{R}^d \rightarrow \mathbb{R}$.
We can then apply the softmax function \emph{within each domain} to produce a soft value assignment:
\begin{equation}
    \varphi^{(t)}(\Cv) = \frac{\exp{\big(o^{(t)}(\Cv)\big)}}{\sum_{\Cv' \in \CV_{X}} \exp{\big(o^{(t)}(\Cv')\big)}}
\end{equation}
This procedure leverages a major strength of our graph construction:
By modeling values as vertices we can directly process arbitrary domains with one GNN.
For larger domains, we simply add more value vertices to the graph.

\subsection{Global Search as an RL Problem}
\label{Section:CSPRL}
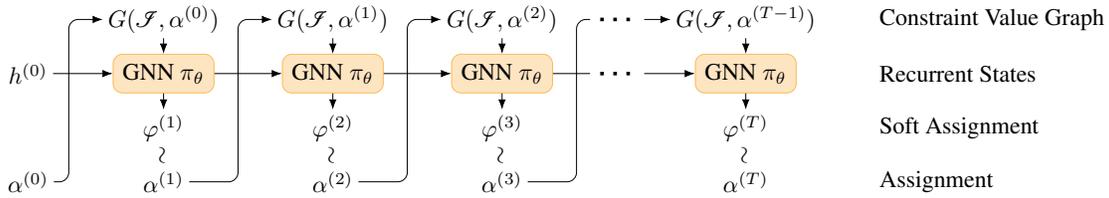
\begin{figure*}[t]
        \centering
        \scalebox{0.9}{
        \begin{tikzpicture}[nodes={inner sep= 2pt}]
    \node at (10.5,0) [anchor=west] {Constraint Value Graph};
    \node at (10.5,-0.8) [anchor=west] {Recurrent States}; 
    \node at (10.5,-1.6) [anchor=west] {Soft Assignment}; 
    \node at (10.5,-2.4) [anchor=west] {Assignment}; 
    
    \node at (-2,-0.8) (h0) { $h^{(0)}$};
    \node at (-2,-2.4) (a0) {$\alpha^{(0)}$};
    
    \node at (0,0) [] (G0) {$G(\mathcal{I}, \alpha^{(0)})$};
    \node at (0,-0.8) [fill=rwth-orange!25, draw=rwth-orange!75, rounded corners=4pt, inner sep = 4pt] (GNN0) {GNN $\pi_\theta$};
    \node at (0,-1.6) [] (phi1) {$\varphi^{(1)}$};
    \node at (0,-2) [] (ti1) {\rotatebox[origin=c]{90}{$\sim$}};
    \node at (0,-2.4) [] (a1) {$\alpha^{(1)}$};
    
    \node at (2.5,0) [] (G1) {$G(\mathcal{I}, \alpha^{(1)})$};
    \node at (2.5,-0.8) [fill=rwth-orange!25, draw=rwth-orange!75, rounded corners=4pt, inner sep = 4pt] (GNN1) {GNN $\pi_\theta$};
    \node at (2.5,-1.6) [] (phi2) {$\varphi^{(2)}$};
    \node at (2.5,-2) [] (ti2) {\rotatebox[origin=c]{90}{$\sim$}};
    \node at (2.5,-2.4) [] (a2) {$\alpha^{(2)}$};
    
    \node at (5,0) [] (G2) {$G(\mathcal{I}, \alpha^{(2)})$};
    \node at (5,-0.8) [fill=rwth-orange!25, draw=rwth-orange!75, rounded corners=4pt, inner sep = 4pt] (GNN2) {GNN $\pi_\theta$};
    \node at (5,-1.6) [] (phi3) {$\varphi^{(3)}$};
    \node at (5,-2) [] (ti3) {\rotatebox[origin=c]{90}{$\sim$}};
    \node at (5,-2.4) [] (a3) {$\alpha^{(3)}$};
    
    \node at (8.6,0) [] (GT) {$G(\mathcal{I}, \alpha^{(T-1)})$};
    \node at (8.6,-0.8) [fill=rwth-orange!25, draw=rwth-orange!75, rounded corners=4pt, inner sep = 4pt] (GNNT) {GNN $\pi_\theta$};
    \node at (8.6,-1.6) [] (phiT) {$\varphi^{(T)}$};
    \node at (8.6,-2) [] (tiT) {\rotatebox[origin=c]{90}{$\sim$}};
    \node at (8.6,-2.4) [] (aT) {$\alpha^{(T)}$};
    
    \draw [-{Latex[length=1.5mm]}] (h0) -- (GNN0);
    \draw [-{Latex[length=1.5mm]}, rounded corners] (a0.east) -- (-1.4,-2.4) -- (-1.4,0) -- (G0.west);
    
    \draw [-{Latex[length=1.5mm]}] (G0) -- (GNN0);
    \draw [-{Latex[length=1.5mm]}] (GNN0) -- (phi1);
    
    \draw [-{Latex[length=1.5mm]}] (GNN0) -- (GNN1);
    \draw [-{Latex[length=1.5mm]}, rounded corners] (a1.east) -- (1.1,-2.4) -- (1.1,0) -- (G1.west);
    
    \draw [-{Latex[length=1.5mm]}] (G1) -- (GNN1);
    \draw [-{Latex[length=1.5mm]}] (GNN1) -- (phi2);
    
    \draw [-{Latex[length=1.5mm]}] (GNN1) --  (GNN2);
    \draw [-{Latex[length=1.5mm]}, rounded corners] (a2.east) -- (3.6,-2.4) -- (3.6,0) -- (G2.west);
    
    \draw [-{Latex[length=1.5mm]}] (G2) -- (GNN2);
    \draw [-{Latex[length=1.5mm]}] (GNN2) -- (phi3);
    
    \draw [] (GNN2) -- (6.3,-0.8);
    \draw [ rounded corners] (a3.east) -- (6.1,-2.4) -- (6.1,0) -- (6.3,0);
    \draw [loosely dotted, line width=1.5pt] (6.45,-0.8) -- (7,-0.8);
    \draw [loosely dotted, line width=1.5pt] (6.45,0) -- (7,0);
    \draw [-{Latex[length=1.5mm]}] (7.1,-0.8) -- (GNNT);
    \draw [-{Latex[length=1.5mm]}] (7.1,0) -- (GT);
    
    \draw [-{Latex[length=1.5mm]}] (GT) -- (GNNT);
    \draw [-{Latex[length=1.5mm]}] (GNNT) -- (phiT);
\end{tikzpicture}
        }
        \caption{Illustration of a run of \method{} on a given CSP instance $\mathcal{I}$. We iteratively apply our policy GNN $\pi_\theta$ to the constraint value graph $G(\mathcal{I}, \alpha^{(t-1)})$ of $\mathcal{I}$ and the current assignment $\alpha^{(t-1)}$. From this we obtain a soft assignment $\varphi^{(t)}$ from which the next assignment $\alpha^{(t)}$ is sampled freely with no restrictions to locality.}
        \label{Fig:Alg}
\end{figure*}
We deploy the policy GNN $\pi_\theta$ as a trainable search heuristic.
Note that a single GNN $\pi_{\theta}$ can search for solutions on any given CSP instance.
\method{} takes a \csp{} instance $\mathcal{I}$ and a parameter $T\in\mathbb{N}$ as input and outputs a sequence $\boldsymbol{\alpha}=\alpha^{(0)},\dots,\alpha^{(T)}$ of assignments for $\mathcal{I}$.
The initial assignment $\alpha^{(0)}$ is simply drawn uniformly at random.
In each iteration $1 \leq t \leq T$ the policy GNN $\pi_\theta$ is applied to the current constraint value graph $G(\mathcal{I}, \alpha^{(t-1)})$ to generate a new soft assignment $\varphi^{(t)}$.
The next assignment $\alpha^{(t)} \sim \varphi^{(t)}$ is then sampled from the predicted soft assignment by drawing a new value $\alpha^{(t)}(X)$ for all variables $X$ independently and in parallel without imposing any restrictions on locality. 
Any number of variables may change their value in each iteration which makes our method a \emph{global} search heuristic.
This allows \method{} to modify different parts of the solution simultaneously to speed up the search.
Figure \ref{Fig:Alg} provides a visual illustration of the overall process.
Formally, our action space is the set of all assignments for the input instance, one of which must be chosen as the next assignment in each iteration $t$.
This set is extremely large for many \csp{}s, with up to $10^{50}$ assignments to choose from for some of our training instances.
Despite this, we found standard policy gradient descent algorithms to be effective and stable during training.
%
\subsubsection{Rewarding Iterative Improvements}
We devise a reward scheme that assigns a real-valued reward $r^{(t)}$ to each generated assignment $\alpha^{(t)}$.
A simple approach would be to use the quality $Q_\mathcal{I}(\alpha^{(t)})$ as a reward.
However, we found that models trained with this reward tend to get stuck in local maxima and have comparatively poor performance.
Intuitively, this simple reward scheme immediately punishes the policy for stepping out of a local maximum causing stagnating behavior.

We, therefore, choose a more sophisticated reward system that avoids this issue.
First, we define the auxiliary variable $q^{(t)} = \max_{t'<t}Q_\mathcal{I}(\alpha^{(t')})$, which tracks the highest quality achieved before iteration $t$.
We then define the reward in iteration $t$ as follows: 
\begin{equation}
        \label{EQ:Reward}
        r^{(t)}=
        \begin{cases} 
            0 & \text{if } Q_\mathcal{I}(\alpha^{(t)}) \leq q^{(t)},\\
            Q_\mathcal{I}(\alpha^{(t)}) - q^{(t)} & \text{if } Q_\mathcal{I}(\alpha^{(t)}) > q^{(t)}.
        \end{cases}
\end{equation}
The policy earns a positive reward in iteration $t$ if the new assignment $\alpha^{(t)}$ satisfies more constraints than any assignment generated in the previous steps.
In this case, the reward is the margin of improvement. 
Note that the reward is $0$ in any step in which the new assignment is not an improvement over the previous best regardless of whether the quality of the solution is increasing or decreasing. 
This reward is designed to encourage $\pi_\theta$ to yield iteratively improving assignments while being agnostic towards how the assignments change between improvements.
Our reward is conceptually similar to that of ECO-DQN \cite{barrett2020exploratory}.
The main difference is that we do not add intermediate rewards for reaching local maxima.
Inductively, we observe that the total reward over all iterations is given by $\sum_{t=1}^T r^{(t)}=q^{(T+1)}-Q_\mathcal{I}(\alpha^{(0)})$. 
For any input instance $\mathcal{I}$ the total reward is maximal (relative to $Q_\mathcal{I}(\alpha^{(0)})$) if and only if the highest achieved quality $q^{(T+1)}$ is the optimal quality for $\CI$.
In Appendix C we provide an ablation study where we compare our reward scheme to the simpler choice of using $Q_\mathcal{I}(\alpha^{(t)})$ directly as a reward.
\subsubsection{Markov Decision Process}
For a given input $\mathcal{I}$ we model the procedure described so far as a Markov Decision Process $\mathcal{M}(\CI)$ which will allow us to deploy standard reinforcement learning methods for training:
The state in iteration $t$ is given by $s^{(t)}=(\alpha^{(t)}, q^{(t)})$ and contains the current assignment and highest quality achieved before step $t$.
The initial assignment $\alpha^{(0)}$ is drawn uniformly at random and $q^{(0)} = 0$.
The space of actions $\mathcal{A}$ is simply the set of all possible assignments for $\mathcal{I}$.\footnote{Formally, the state and action space also contain the recurrent states $h^{(t)}$ which we omit for clarity.}
The soft assignments produced by the policy $\pi_\theta$ are distributions over this action space.
After the next action is sampled from this distribution, the state transition of the MDP is deterministic and updates the state with the chosen assignment and its quality. \jan{Hier nochmal eingefügt das der Zufall im samplen der Action steckt.}
The reward $r^{(t)}$ at time $t$ is defined as in Equation \ref{EQ:Reward}.
\subsubsection{Training}
During training, we assume that some data generating distribution $\Omega$ of CSP instances are given.
We aim to find a policy that performs well on this distribution of inputs.  
Ideally, we need to find the set of parameters $\theta^*$ which maximizes the expected total reward if we first draw an instance from $\Omega$ and then apply the model to it for $T_\text{train}\in\mathbb{N}$ steps:
\begin{equation}
    \label{EQ:Objective}
    \theta^* = \underset{\theta}{\arg\max} \underset{\substack{\mathcal{I} \sim \Omega \\ \boldsymbol{\alpha} \sim \pi_\theta(\mathcal{I})}}{\mathbf{E}} \Big[\sum_{t=1}^{T_\text{train}} \lambda^{t-1} r^{(t)}\Big]
\end{equation}
The discount factor $\lambda \in (0,1]$ and the number of search iterations during training $T_\text{train}$ are both hyperparameters.
Starting with randomly initialized parameters $\theta$, we utilize REINFORCE \citep{williams1992simple} to train $\pi_\theta$ with stochastic policy gradient ascent.
REINFORCE is a natural choice for training \method{} since its complexity \martin{"computational complexity"$\to$"performance} does not depend on the size of the action space $\mathcal{A}$.
Soft assignments allow us to efficiently sample the next assignment $\alpha \sim \varphi$ and recover its probability $\mathbf{P}(\alpha|\varphi)=\prod_{X} \varphi(\alpha(X))$.
These are the only operations on the action space required for REINFORCE.
Note that we use vanilla REINFORCE without a baseline or critic network and we sample a single trajectory for every training instance.
We found this simple version of the algorithm to be surprisingly robust and effective in our setting. 
Details on how the policy gradients are computed are provided in Appendix A.  

\subsection{Implementation and Hyperparameters}
We implement \method{} in PyTorch \footnote{\url{https://github.com/toenshoff/ANYCSP}}.
The code for relabeling $G(\mathcal{I}, \alpha^{(t)})$ in each iteration $t$ is also fully based on PyTorch and is GPU-compatible. 
We implement generalized sparse matrix multiplication in the COO format in CUDA. 
This helps to increase the memory efficiency and speed of the message passes between values and constraints. 
We plan to publish this extension as a stand-alone software package or merge it with PyTorch Sparse to make it accessible to the broader Graph Learning community.

We choose a hidden dimension of $d=128$ for all experiments.
We train with the Adam optimizer for 500K training steps with a batch size of 25.
Training a model takes between 24 and 48 hours, depending on the data.
During training, we set the upper number of iterations to $T_\text{train}=40$.
During testing, we usually run \method{} with a timeout rather than a fixed upper number of iterations $T$.
All hyperparameters are provided in Appendix A. 

For each training distribution $\Omega$ we implement data loaders that sample new instances on-the-fly in each training step.
With our hyperparameters we therefore train each model on 12.5 Million sampled training instances.
We use fixed subsets of 200 instances sampled from each distribution before training as validation data.
The exact generation procedures for each training distribution are provided in Appendix B. 



\section{Experiments}
\label{Section:Experiments}
\begin{figure}
    \centering
    \includegraphics[width=0.49\textwidth]{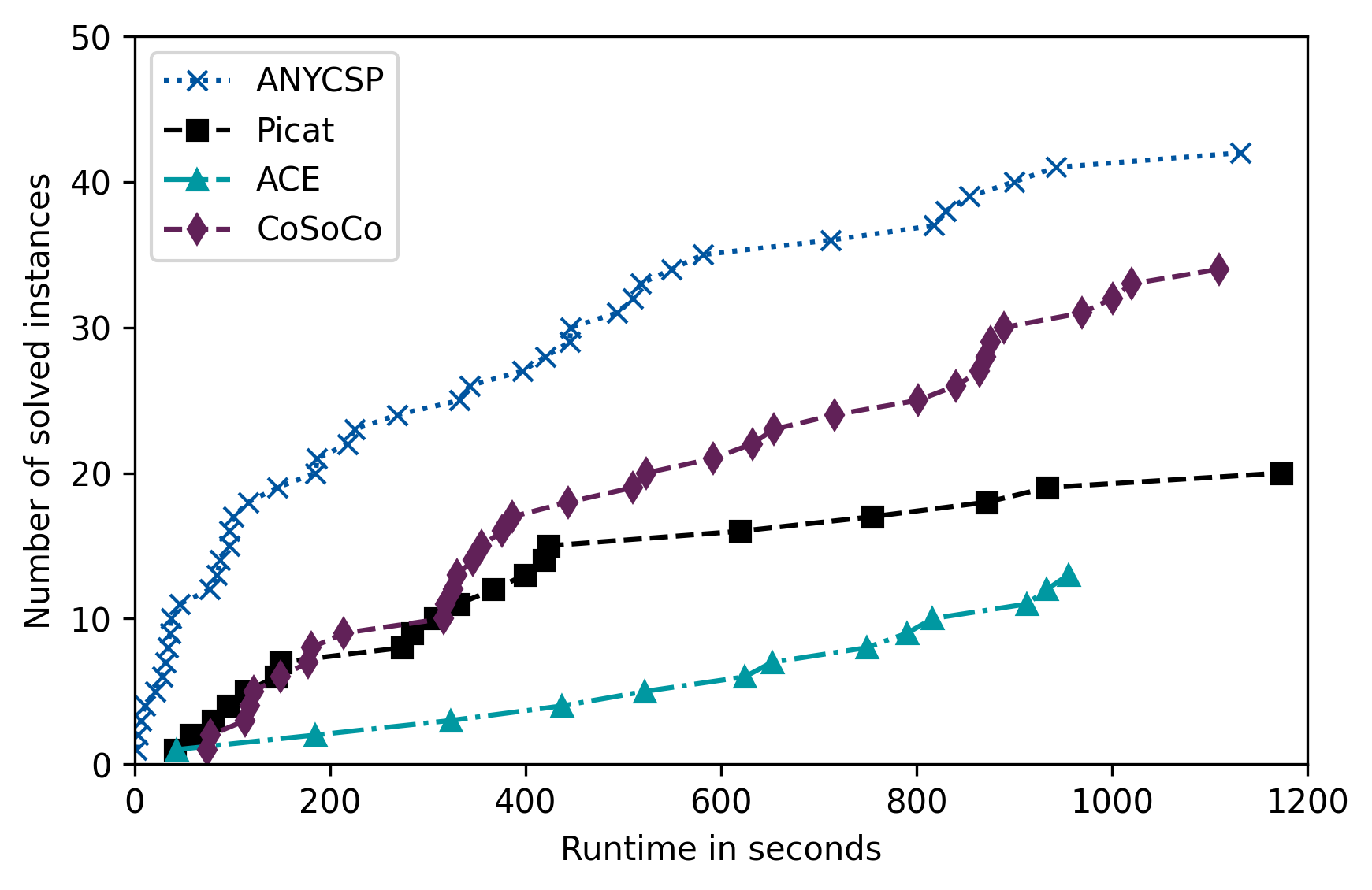}%
    \caption{Survival plot for RB50. The x-axis gives the wall clock runtime in seconds. The y-axis counts the cumulative number of instances solved within a given time.} 
    \label{Fig:RB-Res}
\end{figure}
We evaluate \method{} on a wide range of well-known CSPs: Boolean satisfiability (\threesat{}) and its maximisation version (\msat{} for $k=3,4,5$), graph colorability ($k$-\textsc{Col}), maximum cut (\mcut{}) as well as random CSPs (generated by the so-called \rb{}).
These problems are of high theoretical and practical importance and are commonly used to benchmark CSP heuristics.
We train one \method{} model for each of these problems using randomly generated instances.
Recall that the process of learning problem-specific heuristics with \method{} is purely data-driven as our architecture is generic and can take any CSP instance as input.


We will compare the performance of \method{} to classical solvers and heuristics as well as previous neural approaches.
When applicable, we also tune the configuration of the classical algorithms on our validation data to ensure a fair comparison.
All neural approaches run with one NVIDIA Quadro RTX A6000 GPU with 48GB of memory.
All classical approaches run on an Intel Xeon Platinum 8160 CPU (2.1 GHz) and 64GB of RAM.

\subsubsection{\rb{}}
\begin{table}[t]
\centering
\caption{
    Results on structured Graph Coloring instances. 
    We provide the number of instances solved with a 20 Minute timeout for both splits, each containing 50 instances with chromatic number less than 10 and at least 10, respectively.
}
\label{Tab:KCOL}
\small
\begin{tabular}{ccc}
    \toprule
    \textsc{Method} & $\text{COL}_{<10}$ & $\text{COL}_{\geq10}$ \\
    \midrule
    RUNCSP & 33 & - \\ 
    \textsc{CoSoCo} & 49 & 33 \\ 
    \textsc{Picat} & 49 & 38 \\ 
    \textsc{Greedy} & 16 & 15 \\
    \textsc{DSatur} & 38 & 28 \\
    \textsc{HybridEA} & \textbf{50} & \textbf{40} \\
    \midrule 
    \method{} & \textbf{50} & \textbf{40} \\
    \bottomrule
\end{tabular}

\end{table}
First, we evaluate \method{} on general CSP benchmark instances generated by the \rb{} \cite{xu2003many}.
Our training distribution $\Omega_\text{RB}$ consists of randomly generated \rb{} instances with 30 variables and arity 2.
The test dataset RB50 contains 50 satisfiable instances obtained from the XCSP project \cite{audemard2020xcsp3}.
These instances each contain 50 variables, domains with 23 values and roughly 500 constraints.
They are commonly used as part of the XCSP Competition to evaluate state-of-the-art CSP solvers.
Note that the hardness of \rb{} problems comes from the dense, random constraint relations chosen at the threshold of satisfiability and even instances with 50 variables are very challenging. \jan{Hier noch mehr das RB-Model eingeordnet}
We will compare \method{} to three state-of-the-art CSP solvers: Picat \cite{picat}, ACE \cite{ace} and CoSoCo \cite{audemardcosoco}.
Picat is a SAT-based solver while ACE and CoSoCo are based on constraint propagation.
Picat in particular is the winner of the most recent XCSP Competition \cite{XCSP22}. 
No prior neural baseline exists for this problem.

Figure \ref{Fig:RB-Res} provides a the results on the RB50 dataset.
All algorithms run once on each instance with a 20 Minute timeout.
\method{} solves the most instances by a substantial margin.
The second strongest approach is the CoSoCo solver which solves 34 instances in total, 8 less than \method{}.
Within the timeout of 20 Minutes, \method{} will perform 500K search iterations.
Recall that we set $T_\text{train}=40$.
Therefore, the learned policy generalizes to searches that are over 10K times longer than those seen during training.

\subsubsection{Graph Coloring}
\begin{table}[t]
\centering
\caption{\mcut{} results on Gset graphs. The graphs are grouped by their vertex counts and we provide the mean deviation from the best known cut size.}
\label{Tab:GSET}
\small
\begin{tabular}{crrrr}
    \toprule
    \textsc{Method} & $\tiny|V|\!\!=\!\!800$ & $\tiny|V|\!\!=\!\!1K$ & $\tiny|V|\!\!=\!\!2K$ & $\tiny|V|\!\!\geq\!\!3K\!\!$    \\
    \midrule
    \textsc{Greedy}& 411.44 &359.11 & 737.00 & 774.25\\
    \textsc{SDP}& 245.44 &229.22 &- & -\\
     \textsc{RUNCSP}& 185.89 &156.56 & 357.33 &  401.00 \\
     \textsc{ECO-DQN}& 65.11 &54.67 & 157.00 &  428.25 \\
   \textsc{ECORD}& 8.67 & 8.78& 39.22 & 187.75\\
   \midrule
   $\method{}$ &\textbf{1.22}&\textbf{2.44}&\textbf{13.11}&\textbf{51.63}\\
    \bottomrule
\end{tabular}
\end{table}

We consider the problem of finding a conflict-free vertex coloring given a graph $G$ and number of colors $k$. 
The corresponding CSP instance has variables for each vertex, domains containing the $k$ colors and one binary ``$\neq$''-constraint for each edge.
We train on a distribution $\Omega_\text{COL}$ of graph coloring instances for random graphs with 50 vertices. 
We mix \er{}, \ba{} and random geometric graphs in equal parts.
The number of colors is chosen to be in $[3, 10]$.
As test instances we use 100 structured benchmark graphs with known chromatic number $\mathcal{X}(G)$.
The instances are obtained from a collection of hard coloring instances commonly used to benchmark heuristics\footnote{\url{https://sites.google.com/site/graphcoloring/vertex-coloring}}.
They are highly structured and come from a wide range of synthetic and real problems.
We divide the test graphs into two sets with 50 graphs each:
$\text{COL}_{<10}$ contains graphs with $\mathcal{X}(G) < 10$ and $\text{COL}_{\geq10}$ contains graphs with $\mathcal{X}(G) \geq 10$.
The graphs in $\text{COL}_{\geq10}$ have up to 1K vertices, 19K edges and a chromatic number of up to 73.
This experiment tests generalization to larger domains and more complex structures.

We compare the performance to three problem specific heuristics: a simple greedy algorithm, the classic heuristic DSATUR \cite{brelaz1979new} and the state-of-the-art heuristic HybridEA \cite{galinier1999hybrid}, all implemented efficiently by \citet{lewis2012wide, lewis2015guide}.
We also evaluate the best two CSP solvers from the \rb{} experiment.
The neural baseline RUNCSP is also tested on $\text{COL}_{<10}$.
Unlike \method{}, RUNCSP requires us to fix a domain size before training.
Therefore, we must train one RUNCSP model for each tested chromatic number $4\leq\mathcal{X}(G)\leq9$ and omit testing on $\text{COL}_{\geq10}$.
We use the same training data as \citet{10.3389/frai.2020.580607} for their experiments on structured coloring benchmarks.

Table \ref{Tab:KCOL} provides the number of solved \col{} instances from both splits.
\method{} is on par with HybridEA which solves the most instances of all baselines.
RUNCSP solves significantly fewer instances than \method{} on $\text{COL}_{<10}$ and outperforms only the simple greedy approach.
\method{} solves 40 out of the 50 instances in $\text{COL}_{\geq10}$.
The optimally colored graphs include the largest instance with 73 colors.
Since \method{} trains with 3 to 10 colors the trained model is able to generalize to significantly larger domains. 

\subsubsection{\mcut{}}
For \mcut{} we train on the distribution $\Omega_\text{MCUT}$ of random unweighted \er{} graphs with 100 vertices and an edge probability $p\in[0.05,0.3]$.
Our test data is Gset \cite{Gset}, a collection of commonly used \mcut{} benchmarks of varying structure with 800 to 10K vertices.
We evaluate three neural baselines: RUNCSP, ECO-DQN \cite{barrett2020exploratory} and ECORD \cite{barrett2022learning}.
RUNCSP is also trained on $\Omega_\text{MCUT}$.
We train and validate ECO-DQN and ECORD models with the same data that \citet{barrett2022learning} used for their Gset experiments. 
We omit S2V-DQN \cite{khalil2017learning} since ECO-DQN and ECORD have been shown to yield substantially better cuts.
We adopt the evaluation setup of ECORD and run the neural methods with 20 parallel runs and a timeout of 180s on all unweighted instances of Gset.
The results of a standard greedy construction algorithm and the well-known SDP based approximation algorithm by \citet{goemans1995improved} are also included as classical baselines.
Both are implemented by \citet{CVX}.
SDP runs with a 3 hour timeout for graphs with up to 1K vertices.

Table \ref{Tab:GSET} provides results for Gset.
We divide the test graphs into groups by the number of vertices (8-9 graphs per group) and report the mean deviation from the best-known cuts obtained by \citet{BENLIC20131162} for each method.
\method{} outperforms all 
\martin{"neural" weggelassen} 
baselines across all graph sizes by a large margin.
Recall that RUNCSP trains on a soft relaxation of \mcut{} while ECO-DQN and ECORD are both neural local search approaches.
Neither concept matches the results of our global search approach trained with policy gradients.  

\subsubsection{3-SAT}
\begin{table}[t]
\centering
\caption{Number of solved \threesat{} benchmark instances from SATLIB. For each number of variables there are 100 satisfiable test instances.}
\label{Tab:SATLIB}
\small
\begin{tabular}{cccccc}
    \toprule
    \textsc{Method} & SL50 & SL100 & SL150 & SL200 & SL250 \\
    \midrule
    RLSAT &\textbf{100} & 87 & 67 &27  &12\\
    PDP & 93 & 79 & 72 & 57 & 61 \\
    \textsc{WalkSAT} & \textbf{100} & \textbf{100} & 97 & 93 & 87 \\    \textsc{ProbSAT} & \textbf{100} & \textbf{100} & 97 & 87 & 92 \\
    \midrule 
    $\method{}$ & \textbf{100} & \textbf{100} & \textbf{100} & \textbf{97} & \textbf{99} \\
    \bottomrule
\end{tabular}
\end{table}

For \threesat{} we choose the training distribution $\Omega_\text{3SAT}$ as uniform random \threesat{} instances with 100 variables. 
The ratio of clauses to variables is drawn uniformly from the interval $[4,5]$. 
For \threesat{} we test on commonly used benchmark instances for uniform \threesat{} from SATLIB\footnote{\url{https://www.cs.ubc.ca/~hoos/SATLIB/benchm.html}}.
The test set $\text{SL}N$ contains 100 instances with $N \in \{50, 100, 150, 200, 250\}$ variables each.
The density of these formulas is at the threshold of satisfiability.
We evaluate two neural baselines: RLSAT \citep{yolcu2019learning} and PDP \citep{amizadeh2019pdp}.
PDP is also trained on $\Omega_\text{3SAT}$.
We train RLSAT with the curriculum learning dataset for \threesat{} provided by its authors, since its reward scheme is incompatible with our partially unsatisfiable training instances.
We also adopt the experimental setup of RLSAT, which limits the evaluation run by the number of search steps instead of a timeout.
The provided code for both PDP and RLSAT is comparatively slow and a timeout would compare implementation details rather than the capability of the learned algorithms.
We also evaluate two conventional local search heuristics: The classical WalkSAT algorithm \cite{selman1993local} based on random walks and a modern probabilistic approach called probSAT \cite{probSAT}.
Like \citet{yolcu2019learning}, we apply stochastic boosting and run each method 10 times for 10K steps on every instance.
PDP is deterministic and only applied once to each formula.

Table \ref{Tab:SATLIB} provides the number of solved instances for each tested size.
All compared approaches do reasonably well on small instances with 50 variables.
However, the performance of the two neural baselines drops significantly as the number of variables increases.
\method{} does not suffer from this issue and even outperforms the classical local search algorithms on the three largest instance sizes considered here.

\subsubsection{\textsc{Max}-$\bm{k}$-SAT}
\begin{table}[t]
\centering
\caption{Results on Max-$k$-SAT instances with 10K variables. For each $k\in\{3,4,5\}$ we provide the mean number of unsatisfied clauses over 50 random instances.}
\label{Tab:MAXSAT}
\small
\begin{tabular}{cccc}
    \toprule
    \textsc{Method} & 3CNF & 4CNF & 5CNF \\
    \midrule
    \textsc{WalkSAT} & 2145.28 & 1556.68 & 1685.10 \\
    \textsc{CCLS} & 1567.24 & 1323.14 & 1315.96 \\
    \textsc{SATLike} & 1595.86 & 1188.56 & 1152.88 \\
    \midrule 
    \method{} & \textbf{1537.46} & \textbf{1126.44} & \textbf{1103.14} \\
    \bottomrule
\end{tabular}
\end{table}
We train on the distribution $\Omega_\text{MSAT}$ of uniform random \msat{} instances with 100 variables and $k \in \{3, 4\}$.
Here, the clause/variable ratio is chosen from $[5,8]$ and $[10,16]$ for $k=3$ and $k=4$, respectively.
These formulas are denser than those of $\Omega_\text{3SAT}$ since we aim to train for the maximization task.
Our test data for \msat{} consists of uniform random $k$-CNF formulas generated by us.
For each $k\in\{3,4,5\}$ we generate 50 instances with 10K variables each.
The number of clauses is chosen as 75K for $k=3$, 150K for $k=4$ and 300K for $k=5$.
These formulas are therefore 100 times larger than the training data and aim to test the generalization to significantly larger instances as well as unseen arities, since $k=5$ is not used for training.
Neural baselines for SAT focus primarily on decision problems.
For \msat{} we therefore compare \method{} only to conventional search heuristics:
the classical (Max-)WalkSAT \cite{selman1993local} and two state-of-the-art \textsc{Max-SAT} local search heuristics CCLS \cite{6874523} and SATLike \cite{cai2020old}. 
\begin{figure}
    \centering
    \includegraphics[width=0.49\textwidth]{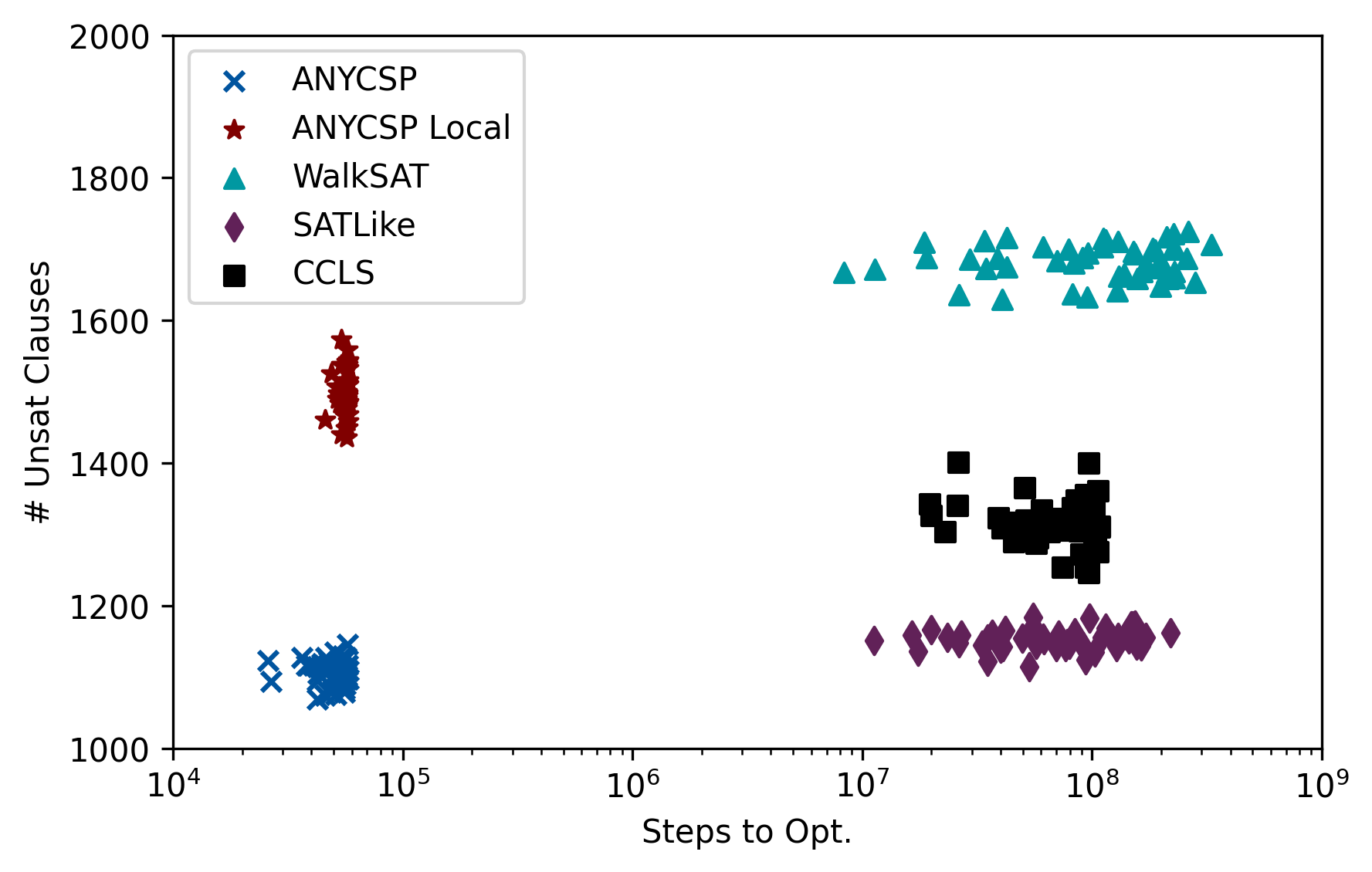}%
    \caption{Detailed results for \textsc{Max}-5-SAT. For each test instance and each method we plot the number of unsatisfied clauses in the best found solution against the search step in which it was found.} 
    \label{Fig:MS-Scatter}
\end{figure}
Table \ref{Tab:MAXSAT} provides a comparison. 
We provide the mean number of unsatisfied clauses after processing each instance with a 20 Minute timeout.
Remarkably, \method{} outperforms all classical baselines by a significant margin. 

We point out that the conventional search heuristics all perform over 100M search steps in the 20 Minute timeout.
\method{} performs less than 100K steps on each instance in this experiment.
The GNN cannot match the speed with which classical algorithms iterate, even though it is accelerated by a GPU.
Despite this, \method{} consistently finds the best solutions.
Figure \ref{Fig:MS-Scatter} evaluates this surprising observation further.
We plot the number of unsatisfied clauses in the best found solution against the search step in which the solution was found (Steps to Opt.) for all methods and all instances of our \textsc{Max}-5-SAT test data.
We also provide the results of a modified \method{} version (\method{} Local defined in Appendix C) that is only allowed to change one variable at a time and is therefore a local search heuristic.
Note that the $x$-axis is logarithmic as there is a clear dichotomy separating neural and classical approaches:
Compared to conventional heuristics \method{} performs roughly three orders of magnitude fewer search steps in the same amount of time.
When restricted to local search, \method{} is unable to overcome this deficit and yields worse results than strong heuristics such as SATLike.
However, when \method{} leverages global search to parallelize refinements across the whole instance it can find solutions in 100K steps that elude state-of-the-art local search heuristics after well over 100M iterations.
\section{Conclusion}
We have introduced \method{}, a novel method for neural combinatorial optimization to learn heuristics for any CSP through a purely data-driven process.
Our experiments demonstrate how the generic architecture of our method can learn effective search algorithms for a wide range of problems.
We also observe that standard policy gradient descent methods like REINFORCE are capable of learning on an exponentially sized action space to obtain global search heuristics for NP-hard problems.
This is a critical advantage when processing large problem instances.

Directions for future work include widening the scope of the architecture even further:
Weighted and partial CSPs are a natural extension of the CSP formalism and could be incorporated through node features and adjustments to the reward scheme.
Variables with real-valued domains may be another viable extension as policy gradient descent is also applicable to infinite continuous action spaces.

\typeout{}
\bibliography{bibliography}

\clearpage
\appendix
\section{Method Details}
\label{Appendix:Method}
Here, we will provide a formal definition of our architecture and training procedure.
We also give information on model selection and hyperparameters and discuss some implementation details.

\subsection{Architecture}


Let us formalize the architecture of our policy GNN $\pi_\theta$.
Recall that the main hyperparameters of $\pi_\theta$ are the latent dimension $d\in\mathbb{N}$ and the aggregation function $\bigoplus$ which we either choose as an element-wise SUM, MEAN or MAX function.
Our GNN is then composed of the following trainable components:
\begin{itemize}
	\item A GRU-Cell $\textbf{G}:\mathbb{R}^d\times\mathbb{R}^d\rightarrow\mathbb{R}^d$ and its trainable initial state $\textbf{h}\in\mathbb{R}^d$.
	This cell is used to update the recurrent value states.
	\item A value encoder MLP $\textbf{E}:\mathbb{R}^{d+1}\rightarrow\mathbb{R}^d$ which merges the information of the recurrent state and the binary label of each value.
	\item Two linear perceptrons $\textbf{M}_\CV,\textbf{M}_\CC:\mathbb{R}^d\rightarrow\mathbb{R}^{2d}$.
	These functions are used to generate the messages that are sent from values to constraints and from constraints to values, respectively. 
	\item Three MLPs $\textbf{U}_\CV,\textbf{U}_\CC,\textbf{U}_\CX:\mathbb{R}^d\rightarrow\mathbb{R}^d$ for combining aggregated messages for values, constraints and variables, respectively.
	\item The output MLP $\textbf{O}:\mathbb{R}^d\rightarrow\mathbb{R}$ which generates the logit scores for each value before we apply the domain-wise softmax.
\end{itemize}
The combined trainable weights of these functions form the parameter vector $\theta$.
The MLPs $\textbf{E}, \textbf{U}_\CV, \textbf{U}_\CC, \textbf{U}_\CX$ and $\textbf{O}$ all have two layers.
The hidden layer is ReLU-activated and has dimension $d$ while the second layer is linear.
We also note that $\textbf{E}, \textbf{M}_\CV, \textbf{M}_\CC, \textbf{U}_\CV,\textbf{U}_\CC$ and $\textbf{U}_\CX$ each apply LayerNorm to their output, which we found to significantly improve convergence during training.

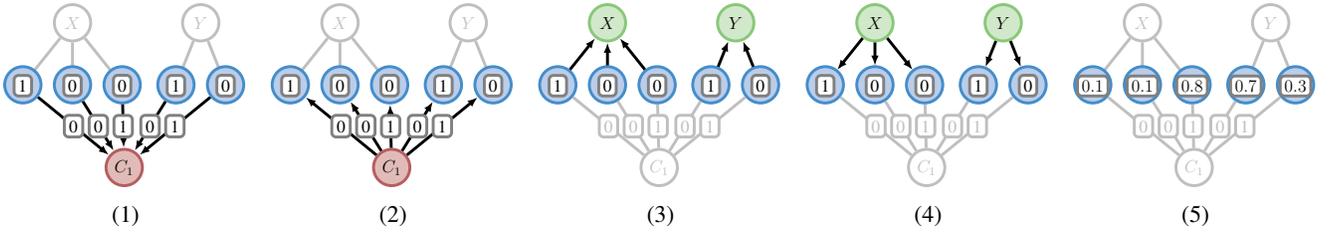
\begin{figure*}[t]
\renewcommand{\thesubfigure}{\arabic{subfigure}}
\centering
\begin{subfigure}{0.2\textwidth}
  \centering
  \scalebox{0.55}{
  \begin{tikzpicture}[
every label/.style = {fill=white, rectangle, draw=black!50, rounded corners=3pt, shift={(-0.0,-0.78)}, minimum width=7pt, inner sep= 3pt, font=\Large},
nodes={
draw,
circle,
minimum width=25pt, 
line width=2pt,
inner sep=1pt,
font=\large}
]
    \node at (0,0) [fill= white, draw= black!25, text = black!25, font = \Large] (X1) {$X$};
    \node at (3.1, 0) [fill= white, draw= black!25, text = black!25, font= \Large] (X2)  {$Y$};
    \draw[line width=2pt, color= black!25] (X1) -- (0,-1.5);
    \draw[line width=2pt, color= black!25] (X1) -- (-1.2, -1.5);
    \draw[line width=2pt, color= black!25] (X1) -- (1.2, -1.5);
    \draw[line width=2pt, color= black!25] (X2) -- (2.5, -1.5);
    \draw[line width=2pt, color= black!25] (X2) -- (3.7, -1.5);
    \node at (-1.2, -1.5) [fill= rwth-blue!25, draw= rwth-blue!70, label={1}] (1)  {}; 
    \node at (0,-1.5) [fill= rwth-blue!25, draw= rwth-blue!70, label={0}] (2)  {}; 
    \node at (1.2, -1.5) [fill= rwth-blue!25, draw= rwth-blue!70, label={0}] (3)  {};
    \node at (2.5, -1.5) [fill= rwth-blue!25, draw= rwth-blue!70, label={1}] (4)  {}; 
    \node at (3.7, -1.5) [fill= rwth-blue!25, draw= rwth-blue!70, label={0}] (5)  {}; 
    \node at (1.25, -3.5) [fill= rwth-carmine!25, draw= rwth-carmine!70, font= \Large] (C1)  {$C_1$};
    \path[every node/.style={fill=white, rectangle, draw=black!50, rounded corners=3pt, font= \Large}, line width=2pt, {Latex[length=2.5mm]}-]
        (C1) edge  node{0} (1)            
        (C1) edge  node{0} (2)
        (C1) edge  node{1} (3)
        (C1) edge  node{0} (4)
        (C1) edge  node{1} (5);
\end{tikzpicture}
  }
  \caption{}
  \label{Fig:MP:2}
\end{subfigure}\hfil
\begin{subfigure}{0.2\textwidth}
  \centering
  \scalebox{0.55}{
  \begin{tikzpicture}[
every label/.style = {fill=white, rectangle, draw=black!50, rounded corners=3pt, shift={(-0.0,-0.78)}, minimum width=7pt, inner sep= 3pt, font=\Large},
nodes={
draw,
circle,
minimum width=25pt, 
line width=2pt,
inner sep=1pt,
font=\large}
]
    \node at (0,0) [fill= white, draw= black!25, text = black!25, font = \Large] (X1) {$X$};
    \node at (3.1, 0) [fill= white, draw= black!25, text = black!25, font= \Large] (X2)  {$Y$};
    \draw[line width=2pt, color= black!25] (X1) -- (0,-1.5);
    \draw[line width=2pt, color= black!25] (X1) -- (-1.2, -1.5);
    \draw[line width=2pt, color= black!25] (X1) -- (1.2, -1.5);
    \draw[line width=2pt, color= black!25] (X2) -- (2.5, -1.5);
    \draw[line width=2pt, color= black!25] (X2) -- (3.7, -1.5);
    \node at (-1.2, -1.5) [fill= rwth-blue!25, draw= rwth-blue!70, label={1}] (1)  {}; 
    \node at (0,-1.5) [fill= rwth-blue!25, draw= rwth-blue!70, label={0}] (2)  {}; 
    \node at (1.2, -1.5) [fill= rwth-blue!25, draw= rwth-blue!70, label={0}] (3)  {};
    \node at (2.5, -1.5) [fill= rwth-blue!25, draw= rwth-blue!70, label={1}] (4)  {}; 
    \node at (3.7, -1.5) [fill= rwth-blue!25, draw= rwth-blue!70, label={0}] (5)  {}; 
    \node at (1.25, -3.5) [fill= rwth-carmine!25, draw= rwth-carmine!70, font= \Large] (C1)  {$C_1$};
    \path[every node/.style={fill=white, rectangle, draw=black!50, rounded corners=3pt, font= \Large}, line width=2pt, -{Latex[length=2.5mm]}]
        (C1) edge  node{0} (1)            
        (C1) edge  node{0} (2)
        (C1) edge  node{1} (3)
        (C1) edge  node{0} (4)
        (C1) edge  node{1} (5);
\end{tikzpicture}
  }
  \caption{}
  \label{Fig:MP:3}
\end{subfigure}\hfil
\begin{subfigure}{0.2\textwidth}
  \centering
  \scalebox{0.55}{
  \begin{tikzpicture}[
every label/.style = {fill=white, rectangle, draw=black!50, rounded corners=3pt, shift={(-0.0,-0.78)}, minimum width=7pt, inner sep= 3pt, font=\Large},
nodes={
draw,
circle,
minimum width=25pt, 
line width=2pt,
inner sep=1pt,
font=\Large}
]
    \node at (0,0) [fill= rwth-green!25, draw= rwth-green!70] (X1) {$X$};
    \node at (3.1, 0) [fill= rwth-green!25, draw=rwth-green!70] (X2)  {$Y$};
    \draw[line width=2pt, {Latex[length=2.5mm]}-] (X1) -- (0,-1.5);
    \draw[line width=2pt, {Latex[length=2.5mm]}-] (X1) -- (-1.2, -1.5);
    \draw[line width=2pt, {Latex[length=2.5mm]}-] (X1) -- (1.2, -1.5);
    \draw[line width=2pt, {Latex[length=2.5mm]}-] (X2) -- (2.5, -1.5);
    \draw[line width=2pt, {Latex[length=2.5mm]}-] (X2) -- (3.7, -1.5);
    \node at (-1.2, -1.5) [fill= rwth-blue!25, draw= rwth-blue!70, label={1}] (1)  {}; 
    \node at (0,-1.5) [fill= rwth-blue!25, draw= rwth-blue!70, label={0}] (2)  {}; 
    \node at (1.2, -1.5) [fill= rwth-blue!25, draw= rwth-blue!70, label={0}] (3)  {};
    \node at (2.5, -1.5) [fill= rwth-blue!25, draw= rwth-blue!70, label={1}] (4)  {}; 
    \node at (3.7, -1.5) [fill= rwth-blue!25, draw= rwth-blue!70, label={0}] (5)  {}; 
    \node at (1.25, -3.5) [fill= white, draw= black!25, text = black!25] (C1)  {$C_1$};
    \path[every node/.style={fill=white, rectangle, draw=black!25, rounded corners=3pt,text= black!25, font= \Large}, line width=2pt,  black!25]
        (C1) edge  node{0} (1)            
        (C1) edge  node{0} (2)
        (C1) edge  node{1} (3)
        (C1) edge  node{0} (4)
        (C1) edge  node{1} (5);
\end{tikzpicture}
  }
  \caption{}
  \label{Fig:MP:4}
\end{subfigure}\hfil
\begin{subfigure}{0.2\textwidth}
  \centering
  \scalebox{0.55}{
  \begin{tikzpicture}[
every label/.style = {fill=white, rectangle, draw=black!50, rounded corners=3pt, shift={(-0.0,-0.78)}, minimum width=7pt, inner sep= 3pt, font=\Large},
nodes={
draw,
circle,
minimum width=25pt, 
line width=2pt,
inner sep=1pt,
font=\Large}
]
    \node at (0,0) [fill= rwth-green!25, draw= rwth-green!70] (X1) {$X$};
    \node at (3.1, 0) [fill= rwth-green!25, draw=rwth-green!70] (X2)  {$Y$};
    \draw[line width=2pt, -{Latex[length=2.5mm]}, shorten >=13pt] (X1) -- (0,-1.5);
    \draw[line width=2pt, -{Latex[length=2.5mm]}, shorten >=13pt] (X1) -- (-1.2, -1.5);
    \draw[line width=2pt, -{Latex[length=2.5mm]}, shorten >=13pt] (X1) -- (1.2, -1.5);
    \draw[line width=2pt, -{Latex[length=2.5mm]}, shorten >=13pt] (X2) -- (2.5, -1.5);
    \draw[line width=2pt, -{Latex[length=2.5mm]}, shorten >=13pt] (X2) -- (3.7, -1.5);
    \node at (-1.2, -1.5) [fill= rwth-blue!25, draw= rwth-blue!70, label={1}] (1)  {}; 
    \node at (0,-1.5) [fill= rwth-blue!25, draw= rwth-blue!70, label={0}] (2)  {}; 
    \node at (1.2, -1.5) [fill= rwth-blue!25, draw= rwth-blue!70, label={0}] (3)  {};
    \node at (2.5, -1.5) [fill= rwth-blue!25, draw= rwth-blue!70, label={1}] (4)  {}; 
    \node at (3.7, -1.5) [fill= rwth-blue!25, draw= rwth-blue!70, label={0}] (5)  {}; 
    \node at (1.25, -3.5) [fill= white, draw= black!25, text = black!25] (C1)  {$C_1$};
    \path[every node/.style={fill=white, rectangle, draw=black!25, rounded corners=3pt,text= black!25, font= \Large}, line width=2pt,  black!25]
        (C1) edge  node{0} (1)            
        (C1) edge  node{0} (2)
        (C1) edge  node{1} (3)
        (C1) edge  node{0} (4)
        (C1) edge  node{1} (5);
\end{tikzpicture}
  }
  \caption{}
  \label{Fig:MP:5}
\end{subfigure}\hfil
\begin{subfigure}{0.2\textwidth}
  \centering
  \scalebox{0.55}{
  \begin{tikzpicture}[
every label/.style = {fill=white, rectangle, draw=black!50, rounded corners=3pt, shift={(-0.0,-0.78)}, minimum width=7pt, inner sep= 3pt, font=\Large},
nodes={
draw,
circle,
minimum width=25pt, 
line width=2pt,
inner sep=1pt,
font=\Large}
]
    \node at (0,0) [fill= white, draw= black!25, text = black!25, font = \Large] (X1) {$X$};
    \node at (3.1, 0) [fill= white, draw= black!25, text = black!25, font= \Large] (X2)  {$Y$};
    \draw[line width=2pt, color= black!25] (X1) -- (0,-1.5);
    \draw[line width=2pt, color= black!25] (X1) -- (-1.2, -1.5);
    \draw[line width=2pt, color= black!25] (X1) -- (1.2, -1.5);
    \draw[line width=2pt, color= black!25] (X2) -- (2.5, -1.5);
    \draw[line width=2pt, color= black!25] (X2) -- (3.7, -1.5);
    \node at (-1.2, -1.5) [fill= rwth-blue!25, draw= rwth-blue!70, label={$0.1$}] (1)  {}; 
    \node at (0,-1.5) [fill= rwth-blue!25, draw= rwth-blue!70, label={$0.1$}] (2)  {}; 
    \node at (1.2, -1.5) [fill= rwth-blue!25, draw= rwth-blue!70, label={$0.8$}] (3)  {};
    \node at (2.5, -1.5) [fill= rwth-blue!25, draw= rwth-blue!70, label={$0.7$}] (4)  {}; 
    \node at (3.7, -1.5) [fill= rwth-blue!25, draw= rwth-blue!70, label={$0.3$}] (5)  {}; 
    \node at (1.25, -3.5) [fill= white, draw= black!25, text = black!25] (C1)  {$C_1$};
    \path[every node/.style={fill=white, rectangle, draw=black!25, rounded corners=3pt,text= black!25, font= \Large}, line width=2pt,  black!25]
        (C1) edge  node{0} (1)            
        (C1) edge  node{0} (2)
        (C1) edge  node{1} (3)
        (C1) edge  node{0} (4)
        (C1) edge  node{1} (5);
\end{tikzpicture}
  }
  \caption{}
  \label{Fig:MP:6}
\end{subfigure}
\caption{Illustration of the message passing scheme in our policy GNN $\pi_\theta$. The process is performed once in each iteration $t$. 
(1) Values pass messages to constraints. (2) Constraints pass messages to values. (3) Values pass messages to variables. (4) Variables pass messages to values. (5) Values predict a new soft assignment.}
\label{Fig:MP}
\end{figure*}

\begin{figure}[H]
\centering
\begin{adjustbox}{minipage=\linewidth,scale=0.75}
  \begin{subfigure}[b]{0.332\linewidth}
    \centering
     \caption*{$t=0$}
    \includegraphics[width=1\linewidth]{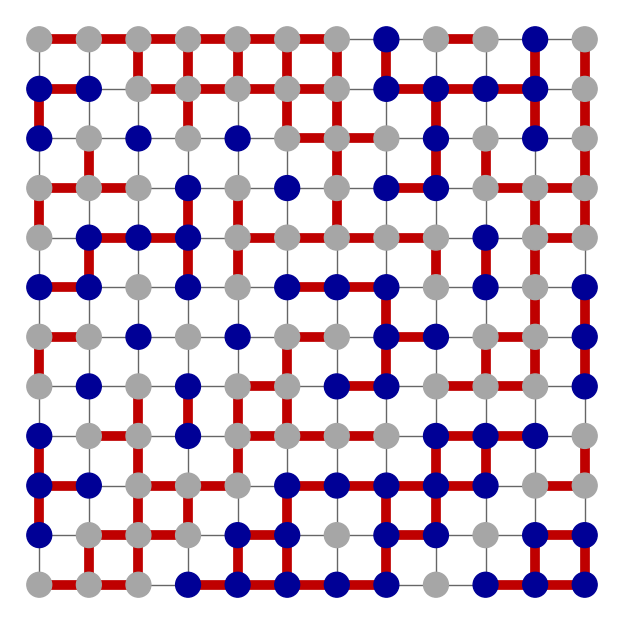}
  \end{subfigure}
  \begin{subfigure}[b]{0.332\linewidth}
    \centering
    \caption*{$t=1$}
    \includegraphics[width=1\linewidth]{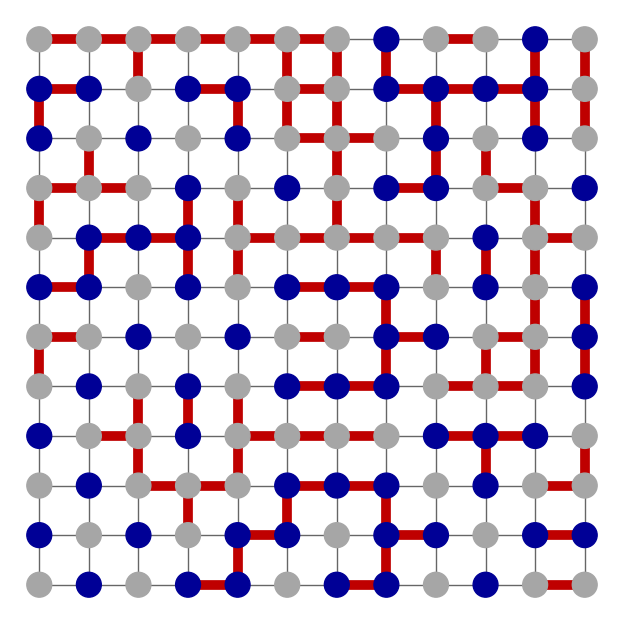}  
  \end{subfigure}
  \begin{subfigure}[b]{0.332\linewidth}
    \centering
    \caption*{$t=2$}
    \includegraphics[width=1\linewidth]{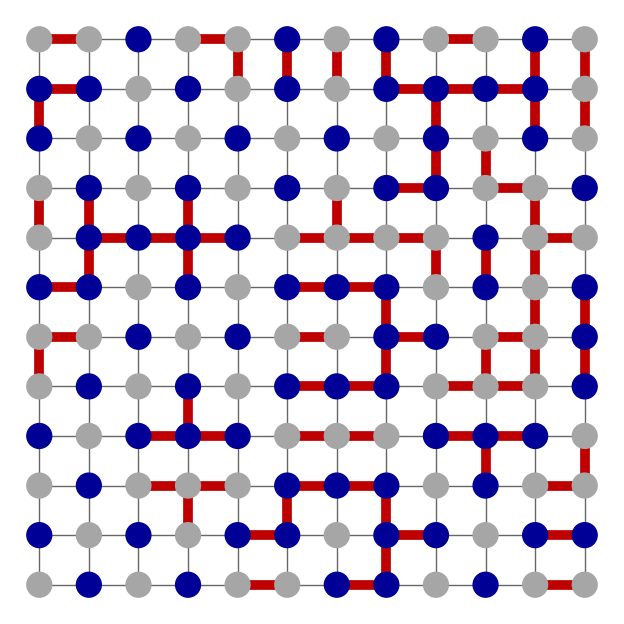} 
  \end{subfigure}
  
  \vspace{0.2cm}
  \begin{subfigure}[b]{0.332\linewidth}
    \centering
    \caption*{$t=3$}
    \includegraphics[width=1\linewidth]{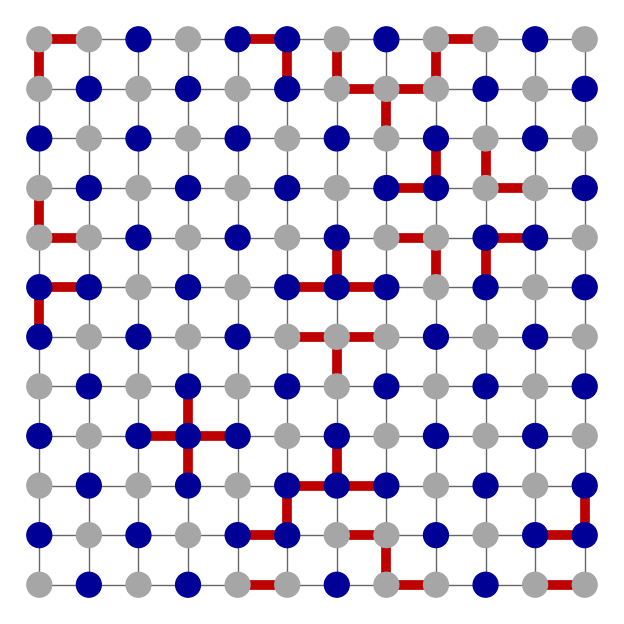}  
  \end{subfigure}
  \begin{subfigure}[b]{0.332\linewidth}
    \centering
    \caption*{$t=4$}
    \includegraphics[width=1\linewidth]{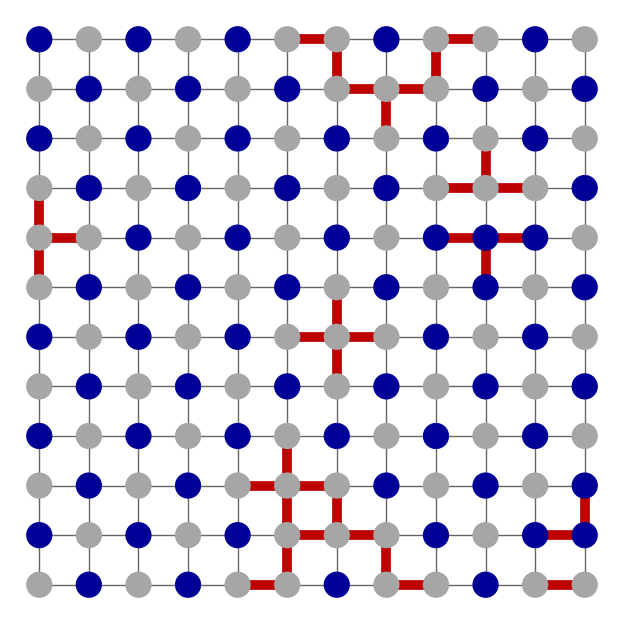} 
  \end{subfigure}
   \begin{subfigure}[b]{0.332\linewidth}
    \centering
    \caption*{$t=5$}
    \includegraphics[width=1\linewidth]{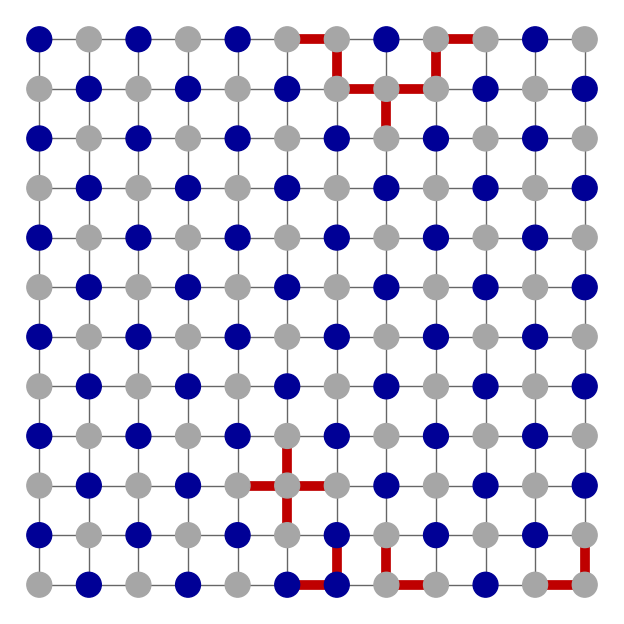}
  \end{subfigure}
   \vspace{0.2cm}
  
  \begin{subfigure}[b]{0.332\linewidth}
    \centering
    \caption*{$t=6$}
    \includegraphics[width=1\linewidth]{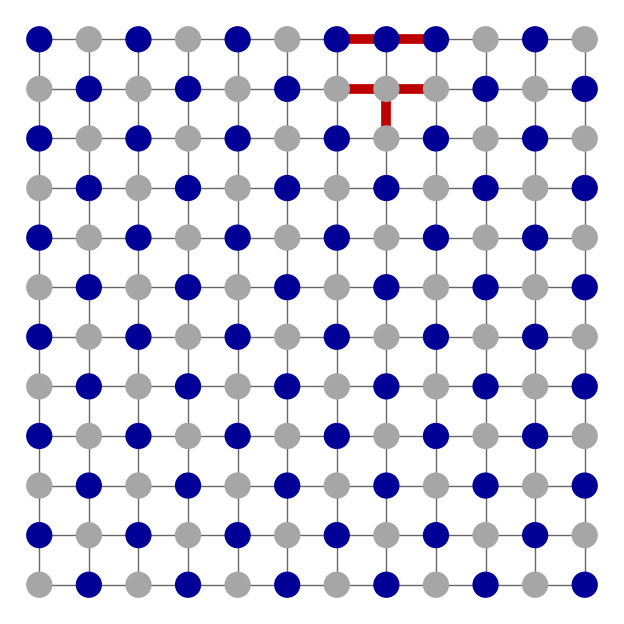}  
  \end{subfigure}
  \begin{subfigure}[b]{0.332\linewidth}
    \centering
    \caption*{$t=7$}
    \includegraphics[width=1\linewidth]{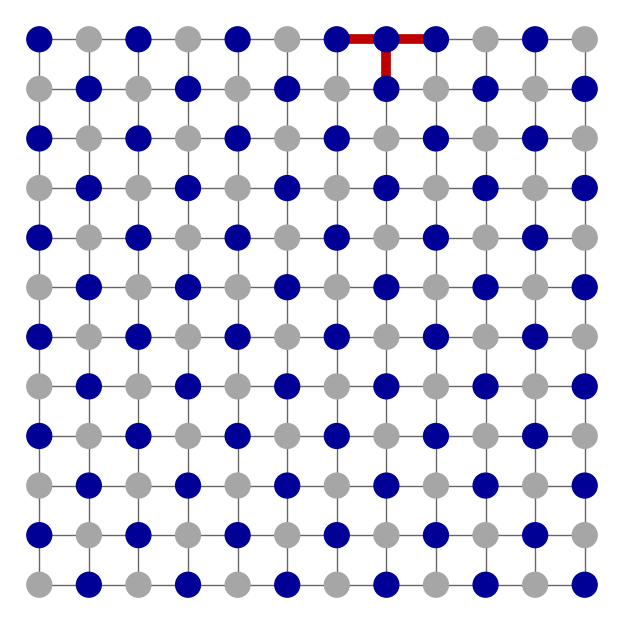} 
  \end{subfigure}
  \begin{subfigure}[b]{0.332\linewidth}
    \centering
    \caption*{$t=8$}
    \includegraphics[width=1\linewidth]{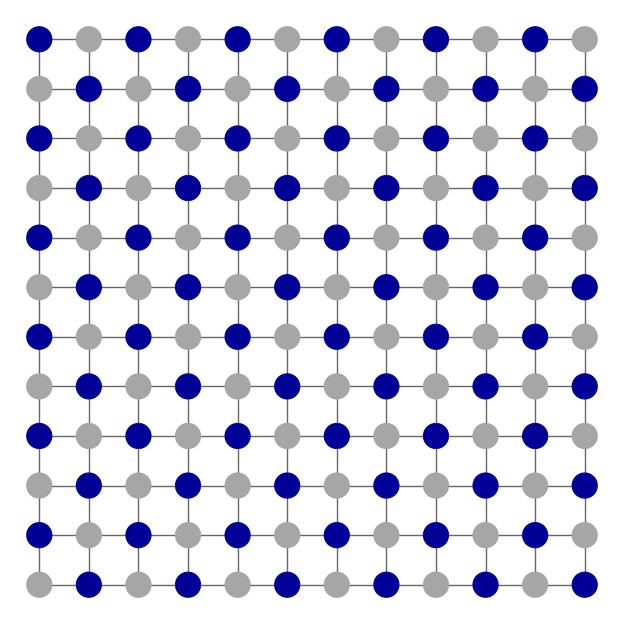}  
  \end{subfigure}
  
  \caption{A 2-coloring for a grid graph with size $12 \times 12$ found by \method{}. Conflicting edges are shown in red.}
  \label{Figure:vis}

 \end{adjustbox}
\end{figure}

In iteration $t$ we associate a recurrent state $h^{(t)}(\Cv)\in\mathbb{R}^d$ with each value $\Cv\in\CV$.
These states are passed on from the previous iteration $t-1$ and initialized as $h^{(0)}(\Cv)=\textbf{h}$.
$\pi_\theta$ then performs the following message passing procedure in each iteration $t$:
First, each value $\Cv\in\CV$ generates a latent state $x^{(t)}(\Cv)$ by applying the encoder $\textbf{E}$ to its recurrent state and its binary label: 
\begin{equation}
	x^{(t)}(\Cv) = \textbf{E}\Big(\big[h^{(t-1)}(\Cv), L_V^{(t-1)}(\Cv)\big]\Big)
\end{equation}
Here, $[\dots]$ denotes concatenation of vectors.
The latent state is then used to generate two messages for each value by applying the message generation MLP $\textbf{M}_\CV$:
\begin{equation}
	m^{(t)}(\Cv,0), m^{(t)}(\Cv,1)  = \textbf{M}_\CV\big(x^{(t)}(\Cv)\big)
\end{equation}
Note that the output of $\textbf{M}_\CV$ has dimension $2d$ and is the stack of both $d$-dimensional messages.
The message $m^{(t)}(\Cv,i)$ is send along all constraint edges $(C, \Cv)$ with label $L_E(C,\Cv)=i$.
Hence, the edge labels are incorporated by generating different messages for each label.
The constraints aggregate these messages and process the result with their message generation function $\textbf{M}_\CC$:
\begin{align}
	&y^{(t)}(C) = \underset{\Cv \in \mathcal{N}(C)}{\bigoplus} m^{(t)}\big(\Cv, L_E(C,\Cv)\big) \\
    &m^{(t)}(C,0), m^{(t)}(C,1) = \textbf{M}_\CC\big(y^{(t)}(C)\big)
\end{align}
These messages are then aggregated by the values that combine the information with their latent state $x$ by applying the update MLP $\textbf{U}_\CV$:
\begin{align}
    y^{(t)}(\Cv) &= \underset{C \in \mathcal{N}(\Cv)\cap\CC}{\bigoplus} \: m^{(t)}(C,L_E(C,\Cv))\\
    z^{(t)}(\Cv) &= \textbf{U}_\CV\big(x^{(t)}(\Cv) + y^{(t)}(v) \big) + x^{(t)}(v)
\end{align}
Note that we added a residual connection around $\textbf{U}_\CV$ for better gradient flow.
In the next phase of our message passing procedure values exchange messages with their respective variables.
To this end, each variable $X\in\CX$ pools the latent states of their respective values and applies $\textbf{U}_\CX$ to obtain a variable-level latent representation $z^{(t)}(X)$:
\begin{equation}
    z^{(t)}(X) = \textbf{U}_\CX\Big(\: \underset{\Cv \in D_X}{\bigoplus} \:  z^{(t)}(\Cv) \:\Big)
\end{equation}
This representation is send back to each value $\Cv \in \CV_X$ of $X$, where it is combined with the value-level latent state by a simple addition.
Note that this final message pass needs no aggregation as every value is connected to exactly one variable.
The result is used as input to the GRU-Cell $\textbf{G}$, which updates the recurrent states of the values:
\begin{equation}
    h^{(t)}(\Cv) = \textbf{G}\Big(h^{(t-1)}(\Cv), z^{(t)}(\Cv) + z^{(t)}(X) \Big)
\end{equation}
Finally, $\pi_\theta$ computes a soft assignment $\varphi^{(t)}$ for $\CI$.
To this end, the MLP $\mathbf{O}$ maps the new recurrent state of each value $\Cv \in \CV_X$ of each variable $X$ to a scalar real number $o^{(t)}(\Cv) = \mathbf{O}(h^{(t)}(\Cv))$.
We can then apply the softmax function \emph{within each domain} to produce a soft value assignment:
\begin{equation}
    \varphi^{(t)}(\Cv) = \frac{\exp{(o^{(t)}(\Cv))}}{\sum_{\Cv' \in \CV_{X}} \exp{(o^{(t)}(\Cv'))}}
\end{equation}
Figure \ref{Fig:MP} provides a visual representation of our message passing procedure.
We also provide the forward pass of \method{} as pseudocode in Algorithm \ref{alg:forward}.
Figure \ref{Figure:vis} visualizes a run of a trained \method{} model on a 2-coloring problem for a grid graph.


\subsection{Training}
Let us formalize how we apply REINFORCE when training an \method{} model.
Recall that our action space is extremely large as we choose one assignment from the set of all possible assignments in each step.
We can handle this action space efficiently because we model probability distributions over this space as soft assignments from which a new value is sampled independently for every variable.
The probability with which a hard assignment $\alpha$ is sampled from a soft assignment $\varphi$ is therefore given by
\begin{equation}
    \label{EQ:prob}
    \mathbf{P}(\alpha|\varphi) = \prod_{X\in\mathcal{X}} \varphi(\alpha(X)).
\end{equation}
Note that sampling one assignment $\alpha \sim \varphi$ and computing its probability according to \ref{EQ:prob} are both efficient operations and are highly parallelizable.
These are the only operation we need on our action space for training and testing.

In each training step, we independently draw a batch of training instances from $\Omega$.
For each such instance $\CI$, we first run \method{} for $T$ steps to generate sequences of soft assignments $\boldsymbol{\varphi}_\theta=\varphi_\theta^{(1)},\dots,\varphi_\theta^{(T)}$ and hard assignments $\boldsymbol{\alpha}=\alpha^{(1)},\dots,\alpha^{(T)}$.
Note that we added $\theta$ as a subscript to the soft assignments to indicate that the parameters in $\theta$ have a partial derivative with respect to the probabilities stored in $\varphi^{(t)}_\theta$.
We first define $G_t$ as the discounted future reward after step $t$:
\begin{equation}
    G_t = \sum_{k=t}^{T}\lambda^{k-t}r^{(k)}
\end{equation}
Here, $\lambda\in(0,1]$ is a discount factor that we usually choose as $\lambda=0.75$.
The purpose of the discount factor is to encourage the policy to earn rewards quickly. 
Our objective is to find parameters $\theta$ that maximize the discounted reward over the whole search:
\begin{equation}
    J(\theta) := \underset{\boldsymbol{\alpha} \sim \pi_\theta(\mathcal{I})}{\mathbf{E}} \Big[\sum_{t=1}^{T} \lambda^{t-1} r^{(t)}\Big]
\end{equation}
REINFORCE \cite{williams1992simple} enables us to estimate the policy gradient as follows:
\begin{align}
    \nabla_\theta J(\theta) &= \nabla_\theta\sum^{T}_{t=1} G_t \log{\mathbf{P}(\alpha^{(t)}|\varphi_\theta^{(t)})}\\
    &= \nabla_\theta\sum^{T}_{t=1}\Big(G_t\!\sum_{X\in\mathcal{X}}\!\log{\big(\varphi_\theta^{(t)}(\alpha^{(t)}(X))\!+\!\epsilon\big)}\!\Big)\label{EQ:loss} 
\end{align}
Equation \ref{EQ:loss} applies Equation \ref{EQ:prob} and adds a small $\epsilon=10^{-5}$ for numerical stability.
These policy gradients are averaged over all instances in the batch and then used for one step of gradient ascent (or rather descent with $-\nabla_\theta J(\theta)$) in the Adam optimizer.
Note that we sample a single trace for each instance in the current batch.
The process is repeated in each training step.
Algorithm \ref{alg:train} provides our overall training procedure as pseudocode.

This training procedure is simply the standard REINFORCE algorithm applied to our Markov Decision Process.
We do not use a baseline or critic network.
We initially expected this simple algorithm to be unable to estimate a useful policy gradient given the unusually large size of our action space and hard nature of our learning problem.
Contrary to this expectation REINFORCE is able to train \method{} effectively.
While more sophisticated RL algorithms have been proposed to address training with large action spaces they are apparently not essential for training with exponentially large action spaces in the context of CSP heuristics.

\subsection{Hyperparameters and Model Selection}
\begin{table*}[t]
\centering
\caption{Selected Hyperparameters for each considered CSP-}
\label{Tab:HypParam}
\begin{tabular}{cccccc}
    \toprule
     & \rb & \col & \threesat & \msat & \mcut \\
    \midrule
    $d$ & 128 & 128 & 128 & 128 & 128 \\
    $\bigoplus$ & MAX & MAX & MAX & MEAN & SUM \\
    $\lambda$ & 0.75 & 0.75 & 0.75 & 0.75 & 0.75 \\
    $T_\text{train}$ & 40 & 40 & 40 & 40 & 40 \\
    batch size & 25 & 25 & 25 & 25 & 25 \\
    lr & $5\cdot10^{-6}$ & $5\cdot10^{-6}$ & $5\cdot10^{-6}$ & $5\cdot10^{-6}$ & $5\cdot10^{-6}$ \\
    \bottomrule
\end{tabular}
\end{table*}
Before training (and hyperparameter tuning) we sample fixed validation datasets of 200 instances from the given distribution of CSP instances.
We usually modify the distribution to yield larger instances than those used for training.
This favors the selection of models that generalize well to larger instances, which is almost always desirable.
Exact details on how the validation distribution differs from the training distribution in each experiment are provided for in Section \ref{Appendix:Results}.
During validation we perform $T_\text{val}=200$ search iterations on each validation instance.
The metric used for selection is the number of unsatisfied constraints in the best solution averaged over all validation instances.
To save compute resources we perform only 100K training steps with each hyperparameter configuration and only perform the full 500K steps of the training with the best configuration.

The aggregation function $\bigoplus \in \{\text{SUM},\text{MEAN},\text{MAX}\}$ is a key hyperparameter.
The choice of $\bigoplus$ is critical for performance, as we observed MAX aggregation to consistently perform best on decision problems but poorly on maximization tasks.
The hidden dimension is set to $d=128$. 
We also validated some models with $d=64$ but larger models seem to be more capable.
We did not increase $d$ further to avoid memory bottlenecks.
We tuned the discount factor $\lambda \in \{0.5, 0.75, 0.9, 0.99\}$ and found the value of $0.75$ to yield the best results in all of our experiments.
The learning rate is initialized as $\text{lr} = 5 \cdot 10^{-6}$ and decays linearly throughout training to a final value of $\text{lr} = 5 \cdot 10^{-7}$.
All model train with a batch size of 25.
We also considered larger batch sizes of 50 and 100 without improvement.
Table \ref{Tab:HypParam} specifies the final configuration used in each experiment.

\subsection{Design Constraints and Bottlenecks}
The primary bottleneck of \method{} is GPU memory.
More specifically, the maximum instance size that can be processed is usually determined by the memory required for the message passes between values and constraints.
Let $\CI=(\CX,\CC,\CD)$ be a CSP instance with constraints of arity $k$ and domains of uniform size $\ell$.
Then the constraint value graph will contain $|\CC| \cdot k \cdot \ell$ constraint edges. 
Constraint edges are represented with a sparse matrix. For each non-zero entry of the sparse matrix (edge), we store the row (outgoing node) and the column (incoming node). Thus the space complexity of storing constraint edges is $\mathcal{O}(|\CC| \cdot k \cdot \ell)$. 
During message passing from values to constraints, each value generates two messages with length d. Those messages are stacked along the first dimension, resulting in a dense matrix with $2 \cdot |\CX| \cdot \ell \cdot d$ entries. To pass messages, we use sparse dense matrix multiplication (SPMM) between the sparse matrix of the edges and the dense matrix of the messages generated from values.
Alternatively, one can also use the scatter operation from the PyTorch Scatter library, but the scatter operation requires the construction of an intermediate tensor stacking the messages send along each edge. 
This allocates extra memory of size $\mathcal{O}(|\CC| \cdot k \cdot \ell \cdot d)$. 
In contrast, SPMM only allocates memory for the result of the aggregation, but no intermediate tensor is build. 
As a result, the space complexity of the SPMM operation is $\mathcal{O}(|\CC|\cdot d)$ for all aggregation types. Combining all of the terms, we get a space complexity of $\mathcal{O}(|\CC| \cdot k \cdot \ell + |\CX| \cdot \ell \cdot d + |\CC| \cdot d)$ for the message passing from values to constraints. The message passing from constraints to values uses the same operations that are used for values to constraints, therefore, the space complexity is the same for both directions. \par
PyTorch Sparse supports generalized SPMM only in CSR format \footnote{\url{https://pytorch-geometric.readthedocs.io/en/latest/notes/sparse\_tensor.html}}. 
Their implementation requires the expensive conversation of the sparse matrix from COO to CSR.
The adjacency matrix between the stack of generated messages and the constraints is re-wired in every iteration $t$ according to the new edge labels.
Converting a new large matrix into CSR format in every step would be prohibitively expensive.
To avoid that, we implement generalized sparse dense matrix multiplication in COO format with CUDA. 
With our newly implemented function we can pass messages memory efficiently and faster.\par
We also experimented with more advanced attention-based aggregation, namely GAT \cite{velivckovic2017graph}. 
However, it did not improve the performance but made the construction of large, intermediate edge-level tensors in the message passes unavoidable.
Due to this, we restrict our focus on the three basic aggregations of element-wise SUM, MEAN and MAX. 

\jan{ToDo...}

\subsection{Relabeling Constraint Value Graphs}
One critical requirement for \method{} is a fast subroutine for recomputing the edge labels $L_E$ given the newly sampled assignment $\alpha^{(t)}$ in step $t$.
Our implementation of this relabeling procedure is based entirely on PyTorch and is GPU accelerated to maximize performance.
Here, we will briefly discuss how this implementation works.

Let $\CI = (\CX,\CD,\CC)$ be a CSP instances and let $\alpha$ be the newly sampled assignment for which we have to compute the edge labels $L_E$.
We first compute the node labels $L_V$, which are a simple binary encoding of $\alpha$.
Let $C \in \CC$ be some constraint with scope $s^C = (X_1,\dots,X_k)$, relation $R^C \in \CD(X_1)\times\dots\times\CD(X_k)$ and arity $k$.
For each tuple $r \in R^C$ we compute a score that counts how many of the values occurring in $t$ are currently chosen by $\alpha$: 
\begin{equation}
    \label{EQ:Label1}   
    s(r) = \sum_{i=1}^{k} L_V(r_i)
\end{equation}
For each value $\Cv \in \CV_{X_i}$ of each variable $X_i$ in the scope of $C$ we then compute the maximum of $s(t)$ over all tuples $r \in R^C$ with $\Cv \in r$:
\begin{equation}
    \label{EQ:Label2}
    m(C, \Cv) = \max_{r \in R^C, \Cv \in r} s(r)
\end{equation}
For each value $\Cv$ we observe that $m(C, \Cv) - L_V(\Cv) = k - 1$ if and only if there exists some tuple $r \in R^C$ with $\Cv \in r$ such that for all $\Cv' \in r, \Cv' \neq \Cv$ we already have $\Cv' \in \alpha$.
This is equivalent to our original definition of $L_E$ and we can use this case distinction to obtain the edge labels: 
\begin{equation}
    \label{EQ:Label3}
    L_E(C,\Cv) =
    \begin{cases} 
        1 & \text{if } m(C, \Cv) - L_V(\Cv) + 1 = k,\\
        0 & \text{otherwise}.
    \end{cases}
\end{equation}

Our implementation maintains edge lists which connect values and constraint edges to their respective tuples across all constraints.
With these edge lists, Equations \ref{EQ:Label1} and \ref{EQ:Label2} are simply scatter operations and Equation \ref{EQ:Label3} is carried out with standard torch arithmetic.
These functions are fully based on the GPU and allow us to rapidly update all edge labels in parallel.
We can optimize this further by allowing relations to be specified in terms of the disallowed tuples (conflicts), rather than the allowed ones.
In this case, we can carry out the exact same procedure, except that we swap the labels in the case distinction of Equation \ref{EQ:Label3}.
This optimization is very useful for many problems.
SAT in particular has constraints that all forbid exactly one tuple.
It is therefore more efficient to work with this one forbidden tuple rather than the $2^k-1$ allowed tuples.

The procedure described here is designed for \emph{extension} constraints, i.e.\ constraints where the relations are defined explicitly with lists of allowed or disallowed tuples.
Many applications of CSPs require constraints for which this is infeasible.
In this case, the relations can only be specified implicitly through \emph{intensions}.
A common example are arithmetic inequalities over discrete numerical domains.
To address this issue, our implementation does provide support for two additional classes of constraints:
\begin{enumerate}
    \item Linear (in)-equalities over numerical domains
    \item ``All-Different''-constraint over many variables
\end{enumerate}
These two types of constraints are commonly found in many CSPs but are hard to express explicitly.
We can still compute the edge labels efficiently on the GPU with similar tricks used for the aforementioned extension constraints.
Since these are not needed to reproduce the results of our main experiments we refer to our source code for further details.
Note that our implementation allows all three supported constraint types to be freely mixed within each instance.

\subsection{Expressiveness}
\jan{Hier Section zu Expressivity eingefügt!}
\martin{Find ich gut, ich hab noch die Referenzen eingefügt.}
A well-known theoretical result on Graph Neural Networks is their correspondence to the Weisfeiler-Lehman isomorphism test.
More specifically, standard message-passing GNNs can not distinguish more structures than the 1-dimensional Weisfeiler-Lehman test.
For node-level tasks this prohibits a GNN to map two different nodes with identical $n$-hop subtrees to different outputs after $n$ message passes.
For some graph structures, such as regular graphs, this makes certain combinatorial tasks, such as graph coloring, fundamentally impossible with standard GNNs.

Crucially, \method{} is \textbf{not} limited by 1-WL.
Our policy GNN $\pi_\theta$ has access to randomness which has been proven to strengthen GNN expressiveness beyond the WL-hierarchy \cite{AbboudCGL21,SatoYK21}.
In each iteration $\pi_\theta$ predicts a soft assignment.
From this we sample a hard assignment and pass it back to the GNN as a binary pattern.
This process can be understood as giving the GNN oracle access to randomness in every iteration.
Soft assignments are not just a way of outputting a new assignment but also provide the means with which $\pi_\theta$ interacts with randomness.
The GNN can learn to predict soft assignments with high variance to break symmetries through random sampling.

Empirically this is also demonstrated in the \mcut{} experiment.
The graphs $G48$ and $G49$ are both 4-regular toroidal graphs.
\method{} computes optimal cuts for both graphs, which correspond to conflict-free 2-colorings.
This could not be achieved by any function limited by 1-WL.

\onecolumn

\begin{algorithm}
\caption{Forward Pass of $\pi_\theta$. All inner for-loops are parallelized.}\label{alg:forward}
\textbf{Input}: \csp{} instance $\CI=(\CX,\CD,\CC)$, Number of steps $T \in \mathbb{N}$.\\
\textbf{Output}: Soft Assignments $\bm{\varphi} = \varphi^{(1)},\dots,\varphi^{(T)}$, Assignments $\bm{\alpha} = \alpha^{(1)},\dots,\alpha^{(T)}$, Rewards $\bm{r} = r^{(1)},\dots,r^{(T)}$

\begin{algorithmic}[1]

\For{$X \in \CX$}
    \State $\alpha^{(0)}(X) \sim \CD(X)$  \Comment{Sample initial assignment uniformly.}
\EndFor
\State $L^{(0)}_V, L^{(0)}_E \leftarrow \text{LABEL}(\CI, \alpha^{(0)})$ \Comment{Get vertex + edge labels.}
\State $q^{(1)} \leftarrow Q_\CI(\alpha^{(0)})\}$ \Comment{Init.\ best prior quality.}
\For{$\Cv \in \CV$}
    \State $h^{(0)}(\Cv) \leftarrow \textbf{h}$  \Comment{\textbf{h} is the learned initial state.}
\EndFor

\For{$t \in \{1,\dots,T\}$}
\For{$\Cv \in \CV$}
\State$x^{(t)}(\Cv) \leftarrow \textbf{E}\Big(\big[h^{(t-1)}(\Cv), L_V^{(t-1)}(\Cv)\big]\Big)$ 
\Comment{Values generate latent state.}

\State 	$m^{(t)}(\Cv,0), m^{(t)}(\Cv,1) \leftarrow \textbf{M}_\CV\big(x^{(t)}(\Cv)\big)$ 
\Comment{Values generate two messages.}
\EndFor

\For{$C \in \CC$}
\State $y^{(t)}(C) = \underset{\Cv \in \mathcal{N}(C)}{\bigoplus} m^{(t)}\big(\Cv, L_E(C,\Cv)\big)$ 
\Comment{Constraints receive messages.}

\State $m^{(t)}(C,0), m^{(t)}(C,1) = \textbf{M}_\CC\big(y^{(t)}(C)\big)$ 
\Comment{Constraints generate messages.}
\EndFor

\For{$\Cv \in \CV$}
\State $y^{(t)}(\Cv) = \underset{C \in \mathcal{N}(\Cv)\cap\CC}{\bigoplus} \: m^{(t)}(C,L_E(C,\Cv))$
\Comment{Values receive messages from constraints.}

\State $z^{(t)}(\Cv) = \textbf{U}_\CV\big(x^{(t)}(\Cv) + y^{(t)}(v) \big) + x^{(t)}(v)$ 
\Comment{Values receive messages.}
\EndFor

\For{$X \in \CX$}
\State $z^{(t)}(X) = \textbf{U}_\CX\Big(\: \underset{\Cv \in D_X}{\bigoplus} \:  z^{(t)}(\Cv) \:\Big)$
\Comment{Variables receive states from values.}
\EndFor

\For{$\Cv \in \CV$}
\State $ h^{(t)}(\Cv) \leftarrow \textbf{G}\Big(h^{(t-1)}(\Cv), z^{(t)}(\Cv) + z^{(t)}(X) \Big)$
\Comment{Update recurrent states.}

\State $o^{(t)}(v) \leftarrow \mathbf{O}(h^{(t)}(v))$
\Comment{Values predict scores.}
\EndFor

\For{$\Cv \in \CV$}
\State $\varphi^{(t)}(\Cv) \leftarrow \frac{\exp{(o^{(t)}(\Cv))}}{\sum_{\Cv' \in \CV(X_\Cv)} \exp{(o^{(t)}(\Cv'))}}$
\Comment{Apply softmax within each domain}
\EndFor

\State $\alpha^{(t)} \sim \varphi^{(t)}$ \Comment{Sample next assignment.}
\State $L^{(t)}_V, L^{(t)}_E \leftarrow \text{LABEL}(\CI, \alpha^{(t)})$ \Comment{Relabel graph.}
\State $r^{(t)} \leftarrow \max\{Q_\CI(\alpha^{(t)}) - q^{(t)}, 0\}$ \Comment{Get Reward.}
\State $q^{(t+1)} \leftarrow \max\{q^{(t)}, Q_\CI(\alpha^{(t)})\}$ \Comment{Update best prior quality.}

\EndFor
\State \Return $\bm{\theta}$, $\bm{\alpha}$, $\bm{r}$ 
\end{algorithmic}
\end{algorithm}

\twocolumn
\onecolumn

\begin{algorithm}
\caption{Training \method{}}\label{alg:train}
\textbf{Input}: Initial parameters $\theta$, training distribution $\Omega$, $\text{train\_steps} \in \mathbb{N}$, $\text{batch\_size} \in \mathbb{N}$, $T_\text{train} \in \mathbb{N}$, $\text{lr} > 0$, $\lambda \in (0, 1]$\\
\textbf{Output}: Trained parameters $\theta$
\begin{algorithmic}[1]
\For{$s \in \{1,\dots,\text{train\_steps}\}$}
    \For{$i \in \{1,\dots,\text{batch\_size}\}$} \Comment{This loop is parallel across all $i$.}
        \State $\CI \sim \Omega$ \Comment{Sample training instance.}
        \State $\bm{\varphi}_\theta, \bm{\alpha}, \mathbf{r} \leftarrow \pi_\theta(\CI, T_\text{train})$ \Comment{Apply policy network.}
        \For{$t \in \{1,\dots,T_\text{train}\}$}
            \State $G_t \leftarrow \sum_{k=t}^{T}\lambda^{k-t}r^{(k)}$
        \EndFor
        \State $\nabla_\theta J_i \leftarrow \nabla_\theta\sum^{T}_{t=1}\Big(G_t\!\sum_{X\in\mathcal{X}} \!\log{\big(\varphi_\theta^{(t)}(\alpha^{(t)}(X))\!+\!\epsilon\big)}\!\Big)$ \Comment{Policy gradient for $i$-th instance in batch.}
    \EndFor
\State $\theta \leftarrow \theta + \frac{\text{lr}}{\text{batch\_size}} \sum_{i} \nabla_\theta J_i$ \Comment{Average gradients and ascent.}
\EndFor
\State \Return $\theta$
\end{algorithmic}
\end{algorithm}

\twocolumn
\section{Experiment Details}
\label{Appendix:Results}
In this section, we will provide additional details on our experimental setup baselines.
We also provide detailed instance-level results for our graph coloring and \mcut{} experiments, since these use structured benchmark instances.
In Table \ref{Tab:software} we provide an overview of all external software used in our experiments.

\subsection{\rb{}}
The \rb{} defines an easy way to generate theoretically hard random CSP instances by randomly choosing a number of disallowed tuples of a fixed arity.

A class of random CSP instances of model RB is denoted RB$(k,n,\alpha,r,p)$ where each instance consists of $n\geq 2$ variables with domain size $d= n^{\alpha}$ for $a > 0$. Each instance has $m = r n  \ln{n}$ constraints for $r > 0$ of arity $k \geq 2$, with each constraint disallowing $t = pd^{k}$ randomly selected tuples. 
Note that the selection of scopes and tuples is performed with repetition. This is due to the fact that the number of repeated constraints and tuples are asymptotically smaller than the total number of constraints and tuples and thus can be neglected.
The hardest of the \rb{} instances occur around the critical value $p_{cr} = 1 - e^{-\alpha/r}$ of $p$  \cite{xu2003many}.

\subsubsection{Data}
Our training data consists of randomly generated \rb{} instances with 30 variables and arity 2.
We randomly select  $d \in (n^{1/k},2n^{1/k}]$ and $m \in [n \log_{k}{d}, 2 n \log_{k}{d}]$ and generate instances with $p = 0,9 p_{cr}$ slightly smaller than the critical values of $p$ to increase the number of satisfiable instances seen during training.
To generate one instance we build $m$ constraints, each by randomly selecting  scope of $k$ distinct variables with repetition and then randomly selecting with repetition a relation of $t$ distinct disallowed tuples.
Our validation data contains 200 instances sampled from the exact same distribution.

The test dataset is obtained from the XCSP project \cite{audemard2020xcsp3} and contains 50 satisfiable \rb{} instances with 50 variables, each with domain size 22 and about 500 constraints of arity 2.
More specifically, we use all instances of the \texttt{Random-RB-2-50-23f} dataset as our test data.

\subsubsection{Baselines}
We used three state-of-the-art CSP-solvers from the XCSP Competition as baselines: Picat \cite{picat}, ACE \cite{ace}, and CoSoCo \cite{audemardcosoco}. Picat is a SAT-based solver and the winner of the most recent XCSP Competition \cite{audemard2020xcsp3}. ACE and CoSoCo are based on constraint propagation. 
We include CoSoCo because it demonstrated very strong performance specifically on binary \rb{} instances in previous CSP Competitions.
Indeed, it also is the best performing baseline in our experiment.

\subsection{Vertex Coloring}
\label{Section:Appendix:Col}
A CSP instance of $k$-COL with the input Graph $G=(V, E)$ has a variable $x_v$ for each vertex $v \in G(V)$, the domain $\CD=\{1,..., k\}$
for each variable, and a constraint $C=((x_v,x_u),R^k_{\neq})$ for each edge $vu \in G(E)$. Here, the Relation $R^k_{\neq}= \{(i,j) | 1 \leq i,j \leq k; i\neq j \}$ implies the color inequality of connected nodes.

We consider the decision problem of \col{}.
That is, we provide the number of colors $k$ as part of the input instance and ask whether or not a conflict-free $k$-coloring exists for the given graph.
If \method{} fails at this task, then it produces a coloring with unsatisfied constraints. 
We do point out that not all of our baselines use this setup.
The greedy heuristic and DSATUR are constructive and yield solutions that are always conflict-free but may have a sub-optimal number of colors.
HybridEA initially constructs a (sub-optimal) conflict-free coloring and then iteratively attempts to lower the number of colors through tabu search and evolutionary optimization.
For all of these methods, we can measure whether or not they produce a conflict-free solution with the optimal number of colors within a given timeout.
However, we should keep these differences in mind during a comparison.

\subsubsection{Data}
To generate training graphs we mix the following 3 distributions uniformly:
\begin{itemize}
    \item \er{} graphs with $n=50$ vertices and edge probability $p \sim U[0.1, 0.3]$
    \item \ba{} graphs with $n=50$ vertices and parameter $m \sim U[2, 10]$
    \item Random geometric graphs with $n=50$ vertices distributed uniformly at random in a 2-dimensional $1 \times 1$ square. The edge threshold radius is drawn uniformly from $r \sim U[0.15,0.3]$.
\end{itemize}
For each graph $G$ drawn from this distribution we then choose a number of colors $k \in [3, 10]$ as follows:
We first apply a linear time greedy coloring heuristic as implemented by NetworkX \cite{hagberg2008exploring} to color the graph without conflict.
If the greedy heuristic required $k'$ colors for $G$, then we pose the problem of coloring $G$ with $k$ colors as the training CSP instance, where $k$ is chosen as:
\begin{equation}
    k = \max\{3, \min\{10, k'-1\}\}
\end{equation}
Intuitively, \method{} has to color each graph with 1 color less than the greedy heuristic.
Some of these instances are unsatisfiable, which is not a problem for our reward scheme and training procedure.
We found this simple method to be very effective at quickly generating graph coloring instances around the threshold of satisfiability with minimal fine-tuning.
The 200 validation instances are generated with the same parameters and procedure, except that we increase the number of vertices for all three graph types to $n=200$.

\subsubsection{Baselines}
RUNCSP was trained on the same data distribution used in its experiments on structured coloring instances (in the Appendix of \citet{10.3389/frai.2020.580607}).
We use a PyTorch implementation of RUNCSP and train each model for a total of 100K steps to ensure convergence.
Recall that RUNCSP requires us to fix one $k$ before training and we train one model for each $k \in \{4,\dots,9\}$.
We consider a graph solved by RUNCSP if the model trained for the graph's chromatic number is able to find a conflict-free coloring.

To evaluate the CSP solvers Picat and CoSoCo in this experiment we reduce each coloring instance to a CSP instance in the XCSP3 format.
As for \method{}, we model the decision variant of graph coloring and fix the known chromatic number of each graph as the domain size.
The solvers then have to find a satisfiable assignment for these instances within the 20-minute timeout.

For HybridEA we optimize the recombination strategy and number of tabu search steps. 
We choose ``nPoint'' for recombination and a factor of 64 for the number of tabu search steps per cycle. 
The greedy algorithm and DSATUR each run in their default configuration.

Table \ref{Tab:KCOL<10} and \ref{Tab:KCOL>10} contain detailed instance-level results of all compared methods for $\text{COL}_{<10}$ and $\text{COL}_{\geq10}$, respectively.

\subsection{\mcut{}}
In our $\mcut{}$ experiment, we only consider instances with positive edge weights.
In this case, \mcut{} is identical to maximum 2-Colorability.
Let $G=(V,E)$ be an input graph.  
We can model the \mcut{} problem for $G$ as a CSP instance by using the same reduction we use for vertex coloring in Section \ref{Section:Appendix:Col} but with the number of colors fixed at $k=2$.
\subsubsection{Data}
Our training distribution $\Omega_\text{MCUT}$ consists if \er{} graphs with $n=100$ vertices and an edge probability sampled uniformly from $p \in [0.05,0.3]$.
Our validation data uses the same distribution for $p$ but with $n=500$ vertices.
\subsubsection{Baselines}
Our classical baselines are a constructive greedy algorithm and a well-known approximation algorithm based on SDP \cite{goemans1995improved}.
For both, we use the implementation of \cite{CVX} and we run SDP with a 3-hour timeout.
We were unable to obtain results for SDP for graphs with over 1000 vertices within this timeout. 

We consider three neural approaches as baselines: RUNCSP \cite{10.3389/frai.2020.580607}, ECORD \cite{barrett2020exploratory} and ECO-DQN \cite{barrett2022learning}. 
RUNCSP is also trained on $\Omega_\text{MCUT}$.
We train ECO-DQN and ECORD on the same data distributions used by \citet{barrett2022learning} in their Gset experiments.
More specifically, ECO-DQN and ECORD train on \er{} graphs with 200 and 500 vertices, respectively.
Both methods were validated with the ``ER500'' validation dataset from their own experiments, which contains \er{} graphs with 500 vertices. 
Note that we select a single model for each method.
\citet{barrett2022learning} suggest selecting different models with different validation datasets modeled after each group of graphs in the Gset test data.
We do not adopt this procedure and select a single model using \er{} graphs for validation.
The goal of our setup is to test the generalization of one model to new sizes and structures not seen during training and validation.

We also experimented with training and validating ECO-DQN and ECORD on our data distributions but found the data chosen by \citet{barrett2022learning} to be better for their methods.
Table \ref{Tab:GSET_extended} provides extended results of the \mcut{} experiment across all used Gset graphs.

\subsection{\threesat{}}
Modeling a Boolean CNF formula $f$ as a CSP instance $\CI=(\CX,\CD,\CC)$ is straightforward.
The set of variables $\CX$ in $\CI$ is simply the set of variables in $f$ and all domains are given by $\CD(X) = \{0,1\}$.
For each clause $c$ in $f$ we add one constraint $C$ to $\CI$ with the same scope of variables as $c$ and the relation $R^C = \{0,1\}^k \setminus \{t_c\}$. 
Here, $k$ is the arity of $c$ and $t_c \in \{0,1\}^k$ is the one combination of values that does now satisfy $c$. 

\subsubsection{Data}
We train on the distribution $\Omega_\text{3SAT}$ of random uniform 3SAT instances with 100 variables and a clause/variable ratio sampled uniformly between $4$ and $5$.
This density is roughly where the threshold of satisfiability for random \threesat{} is.
Many of these instances are unsatisfiable.
For validation, we use formulas with 200 variables and the same density distribution.

\subsubsection{Baselines}
PDP is also trained on $\Omega_\text{3SAT}$.
We used the default PDP configuration for all other options.
RLSAT can not be trained on our distribution, since its reward expects all training instances to be satisfiable.
Furthermore, their training procedure is relatively sensitive and requires a Curriculum Learning to train well.
We, therefore, use the curriculum for \threesat{} provided by the authors of RLSAT. 
Training is performed on a sequence of data sets with increasing variable counts 5, 10, 25, 50 and finally 100. 
This training data is generated by their generators.

We also evaluate the conventional algorithms WalkSAT and probSAT.
For WalkSAT we tuned the ``walk probability'' and ``noise'' parameters but found the default configuration to perform best.
We do run probSAT in its default configuration as well since it has internal heuristics that choose parameters based on the arity of the instance.

We run all methods for 10K steps on each instance. 
We adopt the experimental setting from RLSAT and run each method 10 times on every instance and take the best attempt as the output. 
PDP is deterministic and we run it only once.

\subsection{\textsc{Max}-k-SAT}
For \msat{} we can use the same reduction from Boolean CNF formulas to CSPs that we use for \threesat{}.
The considered formulas are simply denser.

\subsubsection{Data}
Our training distribution $\Omega_\text{MSAT}$ contains $k$-CNF formulas with 100 variables and an arity of $k \in \{3,4\}$.
The clause/variable ratio is sampled uniformly from $[5,8]$ and $[10,16]$ for $k=3$ and $k=4$, respectively.
We validate on formulas with 200 variables and identical density distribution.
No 5-CNF formulas are used for training and model selection.

\subsubsection{Baselines}
In this experiment, we use MaxWalkSAT, a version of the classical WalkSAT algorithm optimized for \textsc{Max}-SAT.
The noise parameter is tuned on our validation data to a value of $10^{-3}$.
The implementations of CCLS and SATLike do not have command line options and run in their default configurations.

\subsection{Cross-Comparison of Trained Models}
\begin{table}[t]
\centering
\caption{Cross-Comparison of training distributions on different test datasets.}
\label{Tab:CrossComp}
\small
\begin{tabular}{cccccc}
    \toprule
     $\Omega$ \!\!\! & RB50 \!\!\! & $\text{COL}_{<10}$ \!\!\! & Gset800 \!\!\! & SL250 \!\!\! & \textsc{Max}-5-CNF \!\!\! \\
    \midrule
    $\Omega_\textbf{RB}$    & \textbf{42} & \textbf{50} & 655.56 & 98 & 6192.18 \\
    $\Omega_\textbf{COL}$   & 15 & \textbf{50} & 868.22 & 96 & 5076.16 \\
    $\Omega_\textbf{MCUT}$  &  0 & 0 & \textbf{1.22} & 0 & 9048.64 \\
    $\Omega_\textbf{3SAT}$  &  0 & 19 & 1213.11 & \textbf{99} & 5001.72 \\
    $\Omega_\textbf{MSAT}$  &  0 & 15 & 1217.67 & 66 & \textbf{1103.14}  \\
    \bottomrule
\end{tabular}

\end{table}
Recall that every \method{} model can take any CSP instance as input.
We train on specific distributions of CSPs to obtain problem specific heuristics.
However, a model trained on graph coloring instances can still process \threesat{} formulas and vice versa.
Naturally, we expect each model to perform best on the distribution it is trained on, but the universality of our architecture does raise interesting questions of how well models trained on one CSP perform on an entirely different CSP.
In this section we aim to study this transferability of learned heuristics across different CSPs.

Table \ref{Tab:CrossComp} compares all models from our main experiments on each other's test data.
For each test dataset, we use the same evaluation metric as the original experiments.
First of all, each model does indeed achieve the best results on the test data of the CSP used for training.
However, the degree to which each model generalizes to other problems varies substantially.
There seems to be a significant compatibility between \rb{} and graph coloring.
The model trained on \rb{} instances solves all coloring instances in $\text{COL}_{<10}$ while the model trained on graph coloring problems solves 15 \rb{} benchmarks instances.
Both of these models also perform very well on decision \threesat{} instances.
Curiously, they even outperform the model trained for \msat{} on decision \threesat{}.
The training distributions $\Omega_{\text{3SAT}}$ and $\Omega_{\text{MSAT}}$ are closely related and one would expected our \msat{} model to do well on \threesat{} as well.
While it does solve 66 of the 100 instances, it does not come close to the 96 and 98 instances solved by the policies trained on $\Omega_{\text{COL}}$ and $\Omega_{\text{RB}}$, respectively.
This surprising observation can most likely be attributed to the choice of the aggregation function.
Like the model trained on $\Omega_{\text{3SAT}}$, the \rb{} and graph coloring policies both use MAX-aggregation.
The \msat{} policy uses MEAN-aggregation.
In this comparison, choosing the right aggregation function appears to be at least as important as training on similar data.
A similar observation also holds for the \mcut{} model.
It performs as well as random guessing on all problems other than \mcut{}.
Note that this model uses SUM-aggregation since this performed best in validation for the \mcut{} problem.
However, SUM-aggregation is not very robust towards changes in the distribution of inputs.
The learned functions are not able to handle larger domains, arities or degrees than those seen during training and the policy is highly specialized towards \mcut{}.
On the other hand, no other model achieves competitive results on \mcut{}, indicating that this problem may require a higher degree of specialization.

Overall, we can conclude that some of our trained models generalize well beyond their training distribution to entirely different CSPs.
However, this is strongly dependent on the specific instance distributions and the robustness of the used aggregation functions.

\onecolumn

\begin{table*}[t]
\centering
\caption{Extended \mcut{} results on Gset graphs. We provide the deviation from the best known cut size.}
\label{Tab:GSET_extended}
\begin{tabular}{ccc|ccccc|cc}
    \toprule
    \textsc{Graph}& $|V|$& $|E|$ & \textsc{Greedy} &\textsc{SDP}&\textsc{RUN-CSP} & \textsc{ECO-DQN} & \textsc{ECORD} &   $\method{}$ \\
    \midrule
    G1 & 800& 19176&675   & 346 &242& 90&\textbf{0} & \textbf{0} \\
    G2& 800   &19176&570 &  343  &226&103 & \textbf{0} & 1 \\
    G3&  800&19176&607  &   326  &207&85 & 10 & \textbf{1} \\
    G4&  800&19176&587  &   330  &222& 55&\textbf{0} & \textbf{0} \\
    G5& 800 &19176&593 &  344  &225& 67 &\textbf{0} & 1 \\
    
    G14& 800& 4694& 164& 145&132& 46 & 15 & \textbf{3} \\
    G15& 800&4661&185 & 126& 146&  46&        16 &    \textbf{0} \\
     G16& 800&4672&158& 128 &141&45 &         20&    \textbf{3} \\
    G17& 800&4667&164& 121 &132&  49&         17&     \textbf{0} \\
    \midrule
    G22& 2000&19990&1019& -&  365&  157&  33  & \textbf{6} \\
    G23& 2000&19990&972 & - &349&181&  21  & \textbf{6} \\
    G24& 2000&19990&965& - &350& 177&  39  & \textbf{9} \\
    G25& 2000&19990&946&- &321& 164 &  32 & \textbf{4} \\
    G26&2000&19990&1001& -&337& 194 &  23  & \textbf{8} \\
    
    G35& 2000&11778&  442& -  &371&  134 &46 & \textbf{19} \\ 
    G36& 2000&11766&  438& -  &358& 141 &53 & \textbf{22} \\ 
    G37& 2000&11785&   415& - & 379& 146&54 & \textbf{22} \\ 
    G38& 2000&11779&  435&  - &386&   119&52 & \textbf{22} \\ 
    \midrule
    G43&1000&9990&473&  213  &156&  66 & \textbf{0} & \textbf{0} \\
    G44& 1000&9990& 532&  294 &143& 57& \textbf{0}& \textbf{0} \\
    G45&1000&9990&463& 329   &135&  52&  \textbf{0} & 1 \\
    G46&1000&9990&461&  264  &159&  69& \textbf{1} & \textbf{1} \\
    G47&1000&9990&455&  289  &158&  33& 6 & \textbf{1} \\
     \midrule
    G48& 3000&6000&\textbf{0}& -  &\textbf{0}&  \textbf{0}& \textbf{0} & \textbf{0} \\
    G49& 3000&6000&\textbf{0}& -   &\textbf{0}& 72 & \textbf{0} & \textbf{0} \\
    G50& 3000&6000&\textbf{0}&   - &2&  32&12& \textbf{0} \\
    \midrule
    G51& 1000&5909& 196&  152  &175&49 &6& \textbf{4} \\
    G52& 1000&5916& 214&  163  &170&  53&21 & \textbf{5} \\
    G53& 1000&5914&  208& 167  &191& 50 &20& \textbf{5} \\
    G54& 1000&5916& 230&  192  &182& 63& 25& \textbf{5} \\
    \midrule
    G55& 5000&12498&1102&  -  &218&405 &211 & \textbf{69} \\
    G58& 5000&29570&1058&  -  &1014 &560&  223& \textbf{56} \\
    \midrule 
    G60&7000&17148&1562&   - &279& 836&    367& \textbf{71} \\
    G63&7000&41459&1384&  -  &1410& 763& 399 & \textbf{74} \\
    \midrule
    G70& 10 000&9999&1088&  - &285&758  &  290 & \textbf{143} \\
    \bottomrule
\end{tabular}
\end{table*}

\begin{table*}[t]
\tiny
\centering
\caption{Extended results on structured Vertex Coloring instances with chromatic number less then 10.}
\label{Tab:KCOL<10}
\resizebox{\textwidth}{!}{\begin{tabular}{crrr|cr|cr|cr|cr|crr|cr|cr}
\toprule
&    & &   &  \multicolumn{2}{c|}{\method{}} &  \multicolumn{2}{c|}{\textsc{RUNCSP}} &   \multicolumn{2}{c|}{\textsc{Picat}}&  
\multicolumn{2}{c|}{\textsc{CoSoCo}}&  
\multicolumn{3}{c|}{\textsc{HybridEA}} &  
\multicolumn{2}{c|}{\textsc{Dsatur}} &
\multicolumn{2}{c}{\textsc{Greedy}}\\
Graph &$|V|$ & $|E|$ & $X(G)$ &  Solved & Time&Solved&Time&Solved&Time&Solved&Time&Solved&Cols&Time&Solved&Cols&Solved&Cols\\
\midrule
   1-FullIns\_3.col &   30 &   100 &  4 &    True &      0.52 &    True &     0.13 &   True &    0.09 &    True &     0.07 &      True &           4 &           0 &    True &         4 &   False &         5 \\
   1-FullIns\_4.col &   93 &   593 &  5 &    True &      0.52 &   False &       TO &   True &     0.1 &    True &     0.07 &      True &           5 &           0 &    True &         5 &   False &         6 \\
   1-FullIns\_5.col &  282 &  3247 &  6 &    True &      0.53 &   False &       TO &   True &    0.19 &    True &     0.11 &      True &           6 &           0 &    True &         6 &   False &         9 \\
1-Insertions\_4.col &   67 &   232 &  5 &    True &      0.51 &    True &     0.13 &   True &    0.08 &    True &     0.01 &      True &           5 &           0 &    True &         5 &    True &         5 \\
1-Insertions\_5.col &  202 &  1227 &  6 &    True &      0.52 &    True &     0.21 &   True &    0.11 &    True &     0.12 &      True &           6 &           0 &    True &         6 &    True &         6 \\
1-Insertions\_6.col &  607 &  6337 &  7 &    True &      0.52 &    True &     3.42 &   True &    0.35 &    True &     0.09 &      True &           7 &           0 &    True &         7 &   False &         8 \\
   2-FullIns\_3.col &   52 &   201 &  5 &    True &      0.51 &    True &     0.13 &   True &    0.08 &    True &     0.01 &      True &           5 &           0 &    True &         5 &    True &         5 \\
   2-FullIns\_4.col &  212 &  1621 &  6 &    True &      0.52 &   False &       TO &   True &    0.16 &    True &     0.03 &      True &           6 &           0 &    True &         6 &   False &         8 \\
   2-FullIns\_5.col &  852 & 12201 &  7 &    True &      0.53 &   False &       TO &   True &    0.73 &    True &     0.19 &      True &           7 &          10 &    True &         7 &   False &        10 \\
2-Insertions\_3.col &   37 &    72 &  4 &    True &      0.51 &    True &     0.14 &   True &    0.07 &    True &     0.04 &      True &           4 &           0 &    True &         4 &    True &         4 \\
2-Insertions\_4.col &  149 &   541 &  5 &    True &      0.52 &    True &     0.15 &   True &    0.08 &    True &     0.02 &      True &           5 &           0 &    True &         5 &   False &         6 \\
2-Insertions\_5.col &  597 &  3936 &  6 &    True &      0.52 &    True &     0.84 &   True &    0.17 &    True &     0.06 &      True &           6 &           0 &    True &         6 &   False &         8 \\
   3-FullIns\_3.col &   80 &   346 &  6 &    True &      0.51 &    True &     0.15 &   True &    0.08 &    True &     0.01 &      True &           6 &           0 &    True &         6 &   False &         7 \\
   3-FullIns\_4.col &  405 &  3524 &  7 &    True &      0.52 &   False &       TO &   True &    0.17 &    True &     0.05 &      True &           7 &           0 &    True &         7 &   False &         8 \\
3-Insertions\_3.col &   56 &   110 &  4 &    True &      0.51 &    True &     0.14 &   True &    0.07 &    True &     0.01 &      True &           4 &           0 &    True &         4 &    True &         4 \\
3-Insertions\_4.col &  281 &  1046 &  5 &    True &      0.52 &    True &     0.17 &   True &     0.1 &    True &     0.02 &      True &           5 &           0 &    True &         5 &    True &         5 \\
3-Insertions\_5.col & 1406 &  9695 &  6 &    True &      0.52 &    True &     7.52 &   True &    0.52 &    True &     0.25 &      True &           6 &          20 &    True &         6 &   False &         8 \\
   4-FullIns\_3.col &  114 &   541 &  7 &    True &      0.52 &    True &     0.26 &   True &    0.12 &    True &     0.01 &      True &           7 &           0 &    True &         7 &    True &         7 \\
   4-FullIns\_4.col &  690 &  6650 &  8 &    True &      0.52 &   False &       TO &   True &    0.31 &    True &     0.09 &      True &           8 &           0 &    True &         8 &   False &        11 \\
4-Insertions\_3.col &   79 &   156 &  4 &    True &      0.51 &    True &     0.14 &   True &    0.06 &    True &     0.01 &      True &           4 &           0 &    True &         4 &   False &         5 \\
4-Insertions\_4.col &  475 &  1795 &  5 &    True &      0.52 &    True &     0.21 &   True &    0.09 &    True &     0.03 &      True &           5 &           0 &    True &         5 &   False &         6 \\
   5-FullIns\_3.col &  154 &   792 &  8 &    True &      0.52 &   False &       TO &   True &    0.08 &    True &     0.02 &      True &           8 &           0 &    True &         8 &    True &         8 \\
   5-FullIns\_4.col & 1085 & 11395 &  9 &    True &      0.51 &   False &       TO &   True &    0.75 &    True &     0.19 &      True &           9 &          20 &    True &         9 &   False &        12 \\
     DSJC125.1.col &  125 &   736 &  5 &    True &      0.95 &   False &       TO &   True &    0.13 &    True &     0.05 &      True &           5 &           0 &   False &         6 &   False &         9 \\
     DSJC250.1.col &  250 &  3218 &  8 &    True &      8.20 &   False &       TO &  False &      TO &   False &       TO &      True &           8 &          30 &   False &        11 &   False &        13 \\
    ash331GPIA.col &  662 &  4181 &  4 &    True &      0.58 &    True &     1.07 &   True &    0.13 &    True &     0.07 &      True &           4 &          10 &   False &         5 &   False &         6 \\
    ash608GPIA.col & 1216 &  7844 &  4 &    True &      0.57 &    True &     1.66 &   True &    0.25 &    True &     0.19 &      True &           4 &          30 &   False &         5 &   False &         6 \\
    ash958GPIA.col & 1916 & 12506 &  4 &    True &      0.62 &   False &       TO &   True &    0.45 &    True &     0.24 &      True &           4 &         110 &   False &         6 &   False &         6 \\
      games120.col &  120 &   638 &  9 &    True &      0.52 &    True &     0.23 &   True &    0.09 &    True &     0.01 &      True &           9 &           0 &    True &         9 &    True &         9 \\
      le450\_5a.col &  450 &  5714 &  5 &    True &     32.04 &   False &       TO &   True &    0.31 &    True &     0.96 &      True &           5 &          80 &   False &        10 &   False &        13 \\
      le450\_5b.col &  450 &  5734 &  5 &    True &      6.64 &   False &       TO &   True &    0.22 &    True &     1.47 &      True &           5 &         110 &   False &        10 &   False &        14 \\
      le450\_5c.col &  450 &  9803 &  5 &    True &      4.02 &    True &        1 &   True &    0.32 &    True &     0.68 &      True &           5 &          40 &   False &        11 &   False &        17 \\
      le450\_5d.col &  450 &  9757 &  5 &    True &      0.74 &    True &      1.1 &   True &    0.33 &    True &     0.15 &      True &           5 &          30 &   False &        12 &   False &        16 \\
      miles250.col &  128 &   387 &  8 &    True &      0.52 &    True &     0.18 &   True &    0.08 &    True &     0.36 &      True &           8 &           0 &    True &         8 &   False &        10 \\
      mug100\_1.col &  100 &   166 &  4 &    True &      0.52 &    True &     0.15 &   True &     1.9 &    True &     0.01 &      True &           4 &           0 &    True &         4 &    True &         4 \\
     mug100\_25.col &  100 &   166 &  4 &    True &      0.52 &    True &     0.15 &   True &    1.85 &    True &     0.55 &      True &           4 &           0 &    True &         4 &    True &         4 \\
       mug88\_1.col &   88 &   146 &  4 &    True &      0.51 &    True &     0.15 &   True &    0.08 &    True &     0.24 &      True &           4 &           0 &    True &         4 &    True &         4 \\
      mug88\_25.col &   88 &   146 &  4 &    True &      0.51 &    True &     0.15 &   True &    0.07 &    True &     0.46 &      True &           4 &           0 &    True &         4 &    True &         4 \\
       myciel3.col &   11 &    20 &  4 &    True &      0.50 &    True &     0.14 &   True &    0.08 &    True &     0.79 &      True &           4 &           0 &    True &         4 &    True &         4 \\
       myciel4.col &   23 &    71 &  5 &    True &      0.51 &    True &     0.14 &   True &    0.07 &    True &     0.68 &      True &           5 &           0 &    True &         5 &    True &         5 \\
       myciel5.col &   47 &   236 &  6 &    True &      0.51 &    True &     0.14 &   True &    0.07 &    True &     0.23 &      True &           6 &           0 &    True &         6 &    True &         6 \\
       myciel6.col &   95 &   755 &  7 &    True &      0.51 &    True &     0.21 &   True &    0.19 &    True &     0.35 &      True &           7 &           0 &    True &         7 &   False &         8 \\
       myciel7.col &  191 &  2360 &  8 &    True &      0.51 &    True &     3.42 &   True &    0.19 &    True &     0.04 &      True &           8 &           0 &    True &         8 &   False &         9 \\
      queen5\_5.col &   25 &   160 &  5 &    True &      0.57 &    True &     0.14 &   True &    0.07 &    True &      0.3 &      True &           5 &           0 &    True &         5 &   False &         7 \\
      queen6\_6.col &   36 &   290 &  7 &    True &      0.57 &   False &       TO &   True &    0.17 &    True &     0.16 &      True &           7 &           0 &   False &         9 &   False &        10 \\
      queen7\_7.col &   49 &   476 &  7 &    True &      3.16 &    True &      0.2 &   True &    0.08 &    True &     0.23 &      True &           7 &           0 &   False &        10 &   False &        12 \\
      queen8\_8.col &   64 &   728 &  9 &    True &      6.35 &   False &       TO &   True &    0.12 &    True &     1.18 &      True &           9 &           0 &   False &        11 &   False &        13 \\
        r125.1.col &  125 &   209 &  5 &    True &      0.52 &    True &     0.14 &   True &    0.08 &    True &     0.07 &      True &           5 &           0 &    True &         5 &   False &         6 \\
        r250.1.col &  250 &   867 &  8 &    True &      0.53 &   False &       TO &   True &     0.1 &    True &     0.08 &      True &           8 &           0 &    True &         8 &   False &         9 \\
   will199GPIA.col &  701 &  6772 &  7 &    True &      0.57 &   False &       TO &   True &    0.26 &    True &      0.1 &      True &           7 &           0 &    True &         7 &   False &        12 \\
\bottomrule
\end{tabular}}
\end{table*}

\begin{table*}[t]
\tiny
\centering
\caption{Extended results on structured Vertex Coloring instances with chromatic number at least 10.
}
\label{Tab:KCOL>10}
\begin{tabular}{crrr|cr|cr|cr|crr|cr|cr}
\toprule
&     &  &   &  \multicolumn{2}{c|}{\method{}} &  \multicolumn{2}{c|}{\textsc{Picat}} &   \multicolumn{2}{c|}{\textsc{CoSoCo}}&  
\multicolumn{3}{c|}{\textsc{HybridEA}}&  
\multicolumn{2}{c|}{\textsc{Dsatur}} &  
\multicolumn{2}{c}{\textsc{Greedy}} \\
Graph &$|V|$ & $|E|$ & $X(G)$ &  Solved & Time&Solved&Time &Solved&Time&Solved&Cols&Time&Solved&Cols&Solved&Cols\\
\midrule
 DSJC125.5.col &  125 &  3891 & 17 &    True &   186.92 &  False &      TO &   False &       TO &      True &          17 &         400 &   False &        22 &   False &        25 \\
   DSJC125.9.col &  125 &  6961 & 44 &    True &     23.4 &   True &   52.65 &   False &       TO &      True &          44 &          40 &   False &        51 &   False &        56 \\
   DSJR500.1.col &  500 &  3555 & 12 &    True &     0.56 &   True &    0.37 &    True &     0.05 &      True &          12 &           0 &    True &        12 &   False &        14 \\
        anna.col &  138 &   493 & 11 &    True &     0.52 &   True &    0.11 &    True &     0.02 &      True &          11 &           0 &    True &        11 &   False &        13 \\
       david.col &   87 &   406 & 11 &    True &     0.51 &   True &     0.1 &    True &     0.02 &      True &          11 &           0 &    True &        11 &   False &        12 \\
flat300\_28\_0.col &  300 & 21695 & 28 &   False &       TO &  False &      TO &   False &       TO &     False &          31 &       TO &   False &        42 &   False &        46 \\
  fpsol2.i.1.col &  496 & 11654 & 65 &    True &     1.14 &   True &   37.88 &    True &     0.35 &      True &          65 &           0 &    True &        65 &    True &        65 \\
  fpsol2.i.2.col &  451 &  8691 & 30 &    True &     0.67 &   True &    5.67 &    True &     0.14 &      True &          30 &           0 &    True &        30 &    True &        30 \\
  fpsol2.i.3.col &  425 &  8688 & 30 &    True &     0.67 &   True &    5.71 &    True &     0.14 &      True &          30 &           0 &    True &        30 &    True &        30 \\
       homer.col &  561 &  1628 & 13 &    True &     0.51 &   True &    0.17 &   False &     TO &      True &          13 &           0 &    True &        13 &   False &        14 \\
        huck.col &   74 &   301 & 11 &    True &     0.51 &   True &    0.09 &    True &     0.01 &      True &          11 &           0 &    True &        11 &    True &        11 \\
  inithx.i.1.col &  864 & 18707 & 54 &    True &     1.07 &   True &   52.22 &    True &     0.41 &      True &          54 &           0 &    True &        54 &    True &        54 \\
  inithx.i.2.col &  645 & 13979 & 31 &    True &     0.62 &   True &   12.77 &    True &     0.22 &      True &          31 &           0 &    True &        31 &    True &        31 \\
  inithx.i.3.col &  621 & 13969 & 31 &    True &     0.66 &   True &   12.51 &    True &     0.23 &      True &          31 &           0 &    True &        31 &    True &        31 \\
        jean.col &   80 &   254 & 10 &    True &     0.52 &   True &    0.09 &    True &     0.01 &      True &          10 &           0 &    True &        10 &    True &        10 \\
   le450\_15a.col &  450 &  8168 & 15 &    True &     4.21 &   True &    1.39 &    True &  1078.32 &      True &          15 &        2720 &   False &        16 &   False &        21 \\
   le450\_15b.col &  450 &  8169 & 15 &    True &     1.97 &   True &    1.29 &    True &     0.13 &      True &          15 &         100 &   False &        16 &   False &        22 \\
   le450\_15c.col &  450 & 16680 & 15 &    True &    44.86 &  False &      TO &   False &       TO &     False &          16 &        TO &   False &        24 &   False &        29 \\
   le450\_15d.col &  450 & 16750 & 15 &   False &       TO &  False &      TO &   False &       TO &      True &          15 &        3190 &   False &        24 &   False &        33 \\
   le450\_25a.col &  450 &  8260 & 25 &    True &     0.55 &   True &    3.21 &    True &     0.23 &      True &          25 &           0 &    True &        25 &   False &        28 \\
   le450\_25b.col &  450 &  8263 & 25 &    True &     0.55 &   True &    3.25 &    True &      0.2 &      True &          25 &           0 &    True &        25 &   False &        27 \\
   le450\_25c.col &  450 & 17343 & 25 &   False &       TO &  False &      TO &   False &       TO &     False &          26 &       TO &   False &        29 &   False &        34 \\
   le450\_25d.col &  450 & 17425 & 25 &   False &       TO &  False &      TO &   False &  TO &     False &          26 &      TO &   False &        28 &   False &        38 \\
   miles1000.col &  128 &  3216 & 42 &    True &     0.69 &   True &    1.11 &    True &     0.12 &      True &          42 &           0 &    True &        42 &   False &        46 \\
   miles1500.col &  128 &  5198 & 73 &    True &     0.72 &   True &   10.02 &    True &     0.27 &      True &          73 &           0 &    True &        73 &   False &        75 \\
    miles500.col &  128 &  1170 & 20 &    True &     0.52 &   True &    0.19 &    True &     0.02 &      True &          20 &           0 &    True &        20 &   False &        22 \\
    miles750.col &  128 &  2113 & 31 &    True &     0.71 &   True &    0.39 &    True &     0.08 &      True &          31 &           0 &    True &        31 &   False &        33 \\
  mulsol.i.1.col &  197 &  3925 & 49 &    True &     0.61 &   True &    2.07 &    True &      0.1 &      True &          49 &           0 &    True &        49 &    True &        49 \\
  mulsol.i.2.col &  188 &  3885 & 31 &    True &     0.58 &   True &     0.9 &    True &     0.17 &      True &          31 &           0 &    True &        31 &    True &        31 \\
  mulsol.i.3.col &  184 &  3916 & 31 &    True &     0.55 &   True &    0.93 &    True &     0.13 &      True &          31 &           0 &    True &        31 &    True &        31 \\
  mulsol.i.4.col &  185 &  3946 & 31 &    True &     0.57 &   True &    0.92 &    True &     0.07 &      True &          31 &           0 &    True &        31 &    True &        31 \\
  mulsol.i.5.col &  186 &  3973 & 31 &    True &     0.55 &   True &    0.95 &    True &     0.06 &      True &          31 &           0 &    True &        31 &    True &        31 \\
  queen10\_10.col &  100 &  2940 & 11 &    True &    16.95 &   True &   30.73 &    True &   351.24 &      True &          11 &           0 &   False &        14 &   False &        15 \\
  queen11\_11.col &  121 &  3960 & 11 &   False &       TO &  False &      TO &   False &       TO &     False &          12 &           TO &   False &        15 &   False &        16 \\
  queen12\_12.col &  144 &  5192 & 12 &   False &       TO &  False &      TO &   False &       TO &     False &          13 &          TO &   False &        17 &   False &        18 \\
  queen13\_13.col &  169 &  6656 & 13 &   False &       TO &  False &      TO &   False &       TO &     False &          14 &          TO &   False &        18 &   False &        20 \\
  queen14\_14.col &  196 &  4186 & 14 &   False &       TO &  False &      TO &   False &       TO &     False &          15 &        TO &   False &        19 &   False &        20 \\
  queen15\_15.col &  225 &  5180 & 15 &   False &       TO &  False &      TO &   False &       TO &     False &          16 &       TO &   False &        22 &   False &        22 \\
  queen16\_16.col &  256 & 12640 & 16 &   False &       TO &  False &      TO &   False &       TO &     False &          17 &      TO &   False &        22 &   False &        23 \\
   queen8\_12.col &   96 &  1368 & 12 &    True &      0.6 &   True &    0.14 &    True &     0.93 &      True &          12 &           0 &   False &        13 &   False &        16 \\
    queen9\_9.col &   81 &  1056 & 10 &    True &     4.97 &   True &    0.21 &    True &    47.81 &      True &          10 &           0 &   False &        13 &   False &        14 \\
     r1000.1.col & 1000 & 14378 & 20 &    True &     0.73 &   True &   11.11 &    True &     0.28 &      True &          20 &          10 &    True &        20 &   False &        26 \\
     r125.1c.col &  125 &  7501 & 46 &    True &     0.73 &   True &   14.05 &    True &     1.14 &      True &          46 &           0 &    True &        46 &   False &        51 \\
      r125.5.col &  125 &  3838 & 36 &    True &     0.78 &   True &    1.45 &    True &    21.86 &      True &          36 &          30 &   False &        38 &   False &        43 \\
      r250.5.col &  250 & 14849 & 65 &    True &    36.41 &   True &   56.35 &   False &       TO &      True &          65 &      442170 &   False &        68 &   False &        78 \\
     school1.col &  385 & 19095 & 14 &    True &     0.79 &   True &    6.42 &   False &       TO &      True &          14 &           0 &   False &        17 &   False &        44 \\
 school1\_nsh.col &  352 & 14612 & 14 &    True &      0.8 &   True &     2.9 &   False &       TO &      True &          14 &          20 &   False &        27 &   False &        38 \\
  zeroin.i.1.col &  211 &  4100 & 49 &    True &     0.66 &   True &    2.41 &    True &     0.08 &      True &          49 &           0 &    True &        49 &    True &        49 \\
  zeroin.i.2.col &  211 &  3541 & 30 &    True &     0.56 &   True &     0.8 &    True &     0.06 &      True &          30 &           0 &    True &        30 &    True &        30 \\
  zeroin.i.3.col &  206 &  3540 &  3 &    True &     0.54 &   True &    0.78 &    True &     0.05 &      True &          30 &           0 &    True &        30 &   False &        31 \\
\bottomrule
\end{tabular}
\end{table*}
\begin{table*}[t]
\centering
\small
\caption{URLs and versions of used softwares.}
\label{Tab:software}
\resizebox{\textwidth}{!}{\begin{tabular}{ccc}
    \toprule
    Software & Version & URL\\
    \midrule
    \textsc{Picat}&3.26&   \url{http://picat-lang.org/}\\
    \textsc{ACE}&2.0& \url{https://github.com/xcsp3team/ace}\\
    \textsc{CoSoCo}&1.12& \url{https://www.cril.univ-artois.fr/XCSP18/files/cosoco-competition.tgz}  \\
    \textsc{PyCSP3}&2.0& \url{https://github.com/xcsp3team/pycsp3}\\
    \textsc{RUNCSP} & 1.0 & \url{https://github.com/toenshoff/RUNCSP-PyTorch} \\
    Greedy + DSATUR + \textsc{HybridEA} (COL) & 2017 & \url{http://rhydlewis.eu/resources/gCol.zip} \\
    \textsc{ECORD}&commit:3463463& \url{https://github.com/tomdbar/ecord} \\
    \textsc{ECO-DQN} & commit:134df73& \url{https://github.com/tomdbar/eco-dqn/}\\
    Greedy + \textsc{SDP} (\mcut{}) & 0.1.2 & \url{https://github.com/hermish/cvx-graph-algorithms} \\
    \textsc{RLSAT} &commit:292e55c & \url{https://github.com/emreyolcu/sat} \\
    \textsc{PDP} & commit:1fca34d& \url{https://github.com/microsoft/PDP-Solver} \\
    \textsc{WalkSAT} & 56 & \url{https://gitlab.com/HenryKautz/Walksat} \\
    \textsc{ProbSAT} & v2014 & \url{https://github.com/satcompetition/2018/blob/master/solvers/probSAT.zip} \\
    \textsc{MaxWalkSAT} & 2013 & https://github.com/stechu/Maxwalksat \\
    \textsc{SATLike}&3.0&   \url{http://lcs.ios.ac.cn/~caisw/MaxSAT.html}\\
    \textsc{CCLS}&2015&   \url{http://lcs.ios.ac.cn/~caisw/MaxSAT.html}\\
    \bottomrule
\end{tabular}}

\end{table*}
\twocolumn

\section{Ablation}
\label{Appendix:Ablation}
We provide an empirical ablation study for two major design choices of \method{}:
Firstly, we want to study the benefit of our exponentially sized action space when compared to a more conventional local search setting.
Secondly, we aim to validate the reward scheme we constructed in Section \ref{Section:CSPRL}.
To this end, we will evaluate two modified versions of \method{}:
\begin{enumerate}
	\item $\text{ANYCSP}_\text{loc}$: A version of \method{} designed to be a local search heuristic.
	We modify $\pi_\theta$ such that the softmax over the scores in $o^{(t)}(\Cv)$ is not performed separately within each domain but over all values in the disjoint union of domains $\CV$:
	\begin{equation}
		\varphi^{(t+1)}(\Cv) = \frac{\exp{(o^{(t)}(\Cv))}}{\sum_{\Cv' \in \CV} \exp{(o^{(t)}(\Cv'))}}
	\end{equation}
	The output $\varphi^{(t+1)}$ of $\pi_\theta$ in iteration $t+1$ is therefore not a soft assignment but a probability distribution over the disjoint union of domains.
	To obtain a new hard assignment $\alpha^{(t+1)}$ we sample a single value $\Cv\in\CC$ from this distribution and set it as the value for its respective variable $X_\Cv$ with $\Cv \in \CV_{X_\Cv}$:
	\begin{align}
		\Cv &\sim \varphi^{(t+1)} \\
		\alpha^{(t+1)} &= \alpha^{(t)}[X_\Cv \!\!=\! \Cv]
	\end{align}
	All variables other than $X_\Cv$ remain unchanged in iteration $t+1$.
	With this modification $\text{ANYCSP}_\text{loc}$ becomes a local search heuristic that only changes one variable at a time.
    The remaining architecture and training procedure are identical to \method{}, including the reward scheme.
	\item $\text{ANYCSP}_\text{qual}$: A version of \method{} trained by using the quality $Q_\CI(\alpha^{(t)})$ of the current assignment as a reward.
	For this configuration to train well, we found it helpful to use the quality of the initial assignment as a baseline:
	\begin{equation}
		r^{(t)} = Q_\CI(\alpha^{(t)}) - Q_\CI(\alpha^{(0)})
	\end{equation}
	Without the subtractive baseline, we found the training to be very unstable.
	Note that the baseline does not solve the fundamental problem of the reward scheme: A heuristic can not leave a local maximum without being immediately punished for doing so.
	Here, we will study how this proposed issue actually effects performance empirically. 
\end{enumerate}
We perform our ablation experiments on the graph coloring, \mcut{} and \msat{} problems.
For each problem, we train both modifications with the same training data and hyperparameters as \method{}. 
\subsection{Results}

Tables \ref{Tab:KCOL-Ab}, \ref{Tab:MAXCUT-Ab} and \ref{Tab:MAXSAT-Ab} contain the results of our ablation study for \col{}, \mcut{} and \msat{}, respectively.
The metrics in each table are identical to those used in our main experiment.

For graph coloring, $\text{ANYCSP}_\text{qual}$ performs significantly worse than the other two versions of our method.
Compared to our main baselines it only outperforms RUNCSP and the simple greedy approach.
$\text{ANYCSP}_\text{loc}$ actually performs reasonably well, as it only solves four graphs less than \method{} across all 100 test instances.
In this experiment the reward scheme seems to contribute more to the performance than the global search action space.
However, only the combination of both in \method{} yields the best results.

On the \mcut{} problem there is no clear hierarchy between $\text{ANYCSP}_\text{loc}$ and $\text{ANYCSP}_\text{qual}$.
However, both ablation versions perform significantly worse than \method{}.
The same seems to hold on the \msat{} problem.
The two modified versions yield similar results but perform far worse than \method{}.

\begin{figure*}[t]
    \centering
    \includegraphics[width=1.\textwidth]{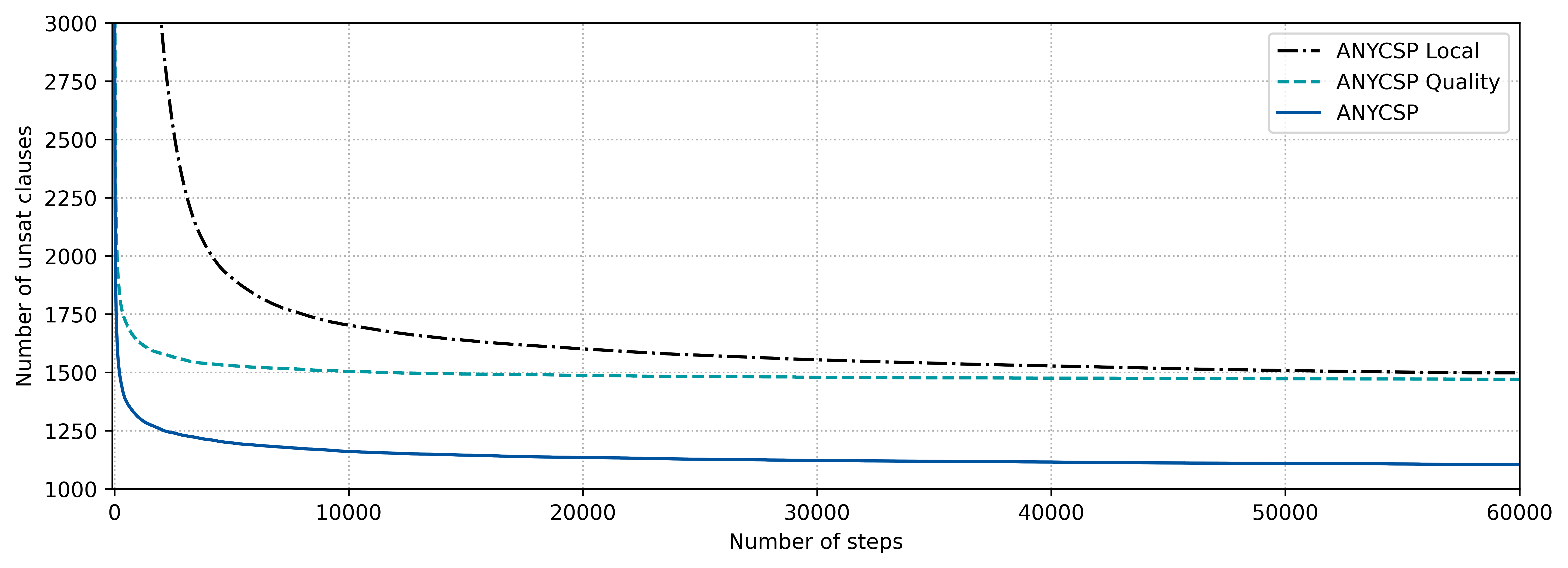}%
    \caption{Number of unsat clauses in the best found solution on our \textsc{Max}-5-SAT test instances. We plot the average over 50 instances as a function in the number of search steps $t$. We compare \method{} and its two ablation versions.} 
    \label{Fig:MS-Ablation}
\end{figure*}

Figure \ref{Fig:MS-Ablation} investigates the differences on the \msat{} problem further.
We plot how the number of unsatisfied clauses in the best found solution evolves throughout the 60K search steps performed by \method{} in 20 minutes.
The curves are averaged over all 50 instances in out \textsc{Max}-5-SAT test data.
Both $\text{ANYCSP}_\text{loc}$ and $\text{ANYCSP}_\text{qual}$ are unable to converge to solutions as good as those found by \method{}, but for different reasons.
$\text{ANYCSP}_\text{loc}$ converges slowly but steadily.
Due to the slow convergence compared to \method{}, it is not able to find equivalent solutions in the same amount of time.
$\text{ANYCSP}_\text{qual}$ initially converges as fast as \method{}.
This is expected, since this version also performs global search and can refine the whole solution in parallel. 
However, it tapers of significantly earlier compared to \method{} and the solution quality remains virtually constant after 20K search steps.
This is the expected problem our reward scheme intends to solve:
During training, $\text{ANYCSP}_\text{qual}$ can not leave local maxima without being punished for doing so by the simple reward scheme.
This inhibits exploration and encourages stagnation.
After 60K steps, $\text{ANYCSP}_\text{loc}$ actually catches up to $\text{ANYCSP}_\text{qual}$ and both ablation versions yield similar results once the 20 minute timeout is reached.

Overall, our experiments and ablation study suggests that our two main design choices are crucial to consistently obtaining strong search heuristics:
\begin{enumerate}
    \item A global search space is necessary to refine the whole solution in parallel and speed up the search.
    Without this advantage, GNN-based heuristics can not compensate for their comparatively high computational cost.
    \item A well-chosen reward scheme that encourages exploration is equally important.
    Without it, global search simply gets stuck faster than local search.
    A simple reward proportional to the quality is not suitable in this regard.
\end{enumerate}
\method{} combines these insights in one generic architecture for all CSPs.
\begin{table}[t]
\centering
\caption{Ablation results on graph coloring.}
\label{Tab:KCOL-Ab}
\small
\begin{tabular}{lcc}
    \toprule
    \textsc{Method} & $\text{COL}_{<10}$ & $\text{COL}_{\geq10}$ \\
    \midrule
    $\method{}_\text{loc}$ & 49 & 37 \\
    $\method{}_\text{qual}$ & 37  & 25 \\
    \method{} & \textbf{50} & \textbf{40} \\
    \bottomrule
\end{tabular}

\end{table}

\onecolumn

\begin{table}[H]
\parbox{.5\linewidth}{
\centering
\small
\caption{Gset results of out ablation models.}
\label{Tab:MAXCUT-Ab}
\begin{tabular}{crrrr}
    \toprule
    \textsc{Method} & $\tiny|V|\!\!=\!\!800$ & $\tiny|V|\!\!=\!\!1K$ & $\tiny|V|\!\!=\!\!2K$ & $\tiny|V|\!\!\geq\!\!3K\!\!$    \\
    \midrule
    $\method{}_\text{loc}$ & 14.22 & 26.00 & 84.00 & 197.50 \\
    $\method{}_\text{qual}$ & 30.11 & 12.89 & 42.56 & 200.75 \\
    $\method{}$ &\textbf{1.22}&\textbf{2.44}&\textbf{13.11}&\textbf{51.63}\\
    \bottomrule
\end{tabular}
}
\hfill
\parbox{.5\linewidth}{
\centering
\small
\caption{Ablation results on Max-$k$-SAT instances.}
\label{Tab:MAXSAT-Ab}
\begin{tabular}{ccc}
    \toprule
    \textsc{Method} & 3CNF & 5CNF \\
    \midrule
    $\method{}_\text{loc}$ & 1697.08 & 1498.46 \\
    $\method{}_\text{qual}$ & 1921.70 & 1471.00 \\
    \method{} & \textbf{1537.46} & \textbf{1103.14} \\
    \bottomrule
\end{tabular}

}
\end{table}

\twocolumn


\end{document}